\documentclass[10pt,twocolumn,letterpaper]{article}
\usepackage[pagenumbers]{cvpr}
\usepackage[utf8]{inputenc}
\usepackage[T1]{fontenc}
\usepackage{nicefrac}
\usepackage{microtype}
\usepackage{algorithm}
\usepackage{algorithmic}
\usepackage{tabularx}
\usepackage{multirow}
\usepackage{array}
\usepackage{ragged2e}
\usepackage{seqsplit}
\usepackage{xurl}
\usepackage{placeins}
\definecolor{cvprblue}{rgb}{0.21,0.49,0.74}
\usepackage[breaklinks,colorlinks,allcolors=cvprblue]{hyperref}
\hypersetup{pdfauthor={},pdftitle={LM-SimBench: A Unified Framework, Dataset, and Benchmark for Large Language Model Training Simulation}}
\title{LM-SimBench: A Unified Framework, Dataset, and Benchmark for Large Language Model Training Simulation}
\author{
Yu Dong$^{1,\dagger}$\quad Hong Li$^{1,\dagger}$\quad Qiang Sun$^{1}$\quad Xu Zhou$^{1}$\\
Jinbiao Huo$^{2}$\quad Qixiu Cheng$^{1,*}$\quad Zhiyuan Liu$^{1,*}$\\[3pt]
{\normalsize $^{1}$Southeast University\quad $^{2}$The Hong Kong Polytechnic University}\\[2pt]
{\small\href{mailto:dongyu_007@seu.edu.cn}{\texttt{dongyu\_007@seu.edu.cn}}\quad
\href{mailto:hongli09@seu.edu.cn}{\texttt{hongli09@seu.edu.cn}}\quad
\href{mailto:220243655@seu.edu.cn}{\texttt{220243655@seu.edu.cn}}\quad
\href{mailto:xuzhou@seu.edu.cn}{\texttt{xuzhou@seu.edu.cn}}}\\
{\small\href{mailto:jinbiao.huo@polyu.edu.hk}{\texttt{jinbiao.huo@polyu.edu.hk}}\quad
\href{mailto:qixiu.cheng@seu.edu.cn}{\texttt{qixiu.cheng@seu.edu.cn}}\quad
\href{mailto:zhiyuanl@seu.edu.cn}{\texttt{zhiyuanl@seu.edu.cn}}}\\[2pt]
{\small $^{\dagger}$These authors contributed equally.}\\
{\small $^{*}$Corresponding authors: Qixiu Cheng and Zhiyuan Liu.}
}
\hypersetup{pdfauthor={Yu Dong, Hong Li, Qiang Sun, Xu Zhou, Jinbiao Huo, Qixiu Cheng, Zhiyuan Liu}}

\begin{document}
\maketitle
\begin{abstract}
  Evaluating candidate parallelization strategies through distributed training
  runs makes strategy selection prohibitively expensive at scale. Performance
  simulation offers a lower-cost alternative, yet controlled comparisons among
  existing simulators remain difficult because they differ in the training
  stages they model and the workloads on which they are evaluated. We
  introduce LM-SimBench\footnote{Code: \href{https://github.com/HongriJiujiu/LM-SimBench}{GitHub}; dataset: \href{https://huggingface.co/datasets/anonymous-ljasd/LM-SimBench}{Hugging Face}.}, a unified framework, profiling
  dataset,
  and benchmark
  for evaluating LLM training simulators. The framework decomposes simulation
  into functional stages from model compilation to schedule-level
  aggregation, connecting execution representations and cost estimators
  to measurable evaluation targets. Following this framework, we profile 23
  model settings from 10 dense and Mixture-of-Experts (MoE) LLM families on an
  Ascend Neural Processing Unit (NPU) cluster. The resulting dataset contains
  348 feasible configurations that jointly vary five parallelism dimensions
  and activation recomputation, with six views spanning configuration, rank,
  operator, event, and kernel granularity. Evaluations of representative
  training simulators demonstrate how LM-SimBench supports comparisons of
  prediction accuracy, assessment of configuration selection, and diagnosis
  of prediction errors. Matched GPU--NPU experiments validate the
  cross-platform applicability of the collection and evaluation protocol
  and demonstrate NPU-guided GPU configuration selection.
\end{abstract}

\section{Introduction}
\label{sec:introduction}

\begin{figure*}[t]
    \centering
    \includegraphics[width=\linewidth]{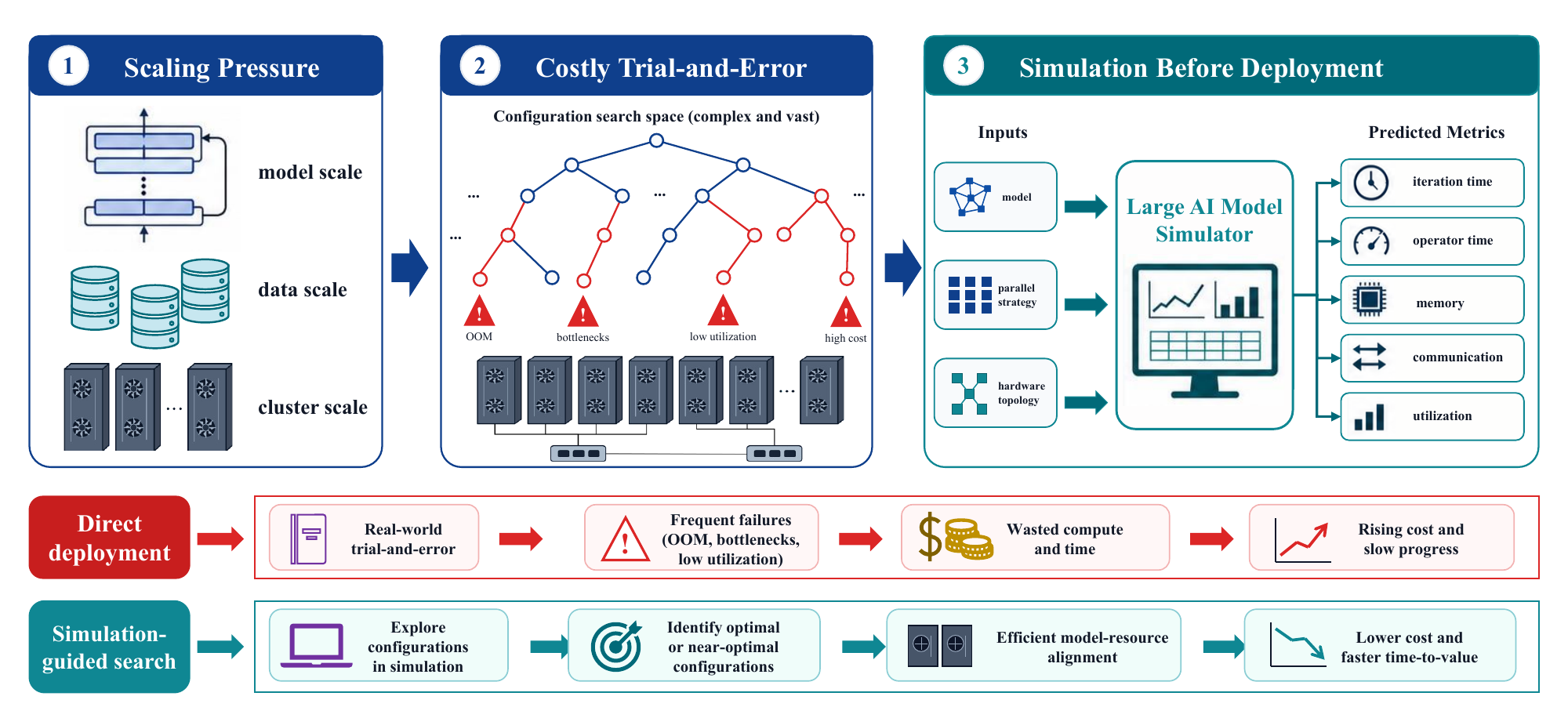}
    \caption{Simulation for model--resource alignment. Effective alignment
    improves utilization and reduces cost; poor alignment creates bottlenecks
    and resource waste. Simulation estimates these effects before deployment,
    reducing the need for repeated full-scale trials.}
    \label{fig:model_resource_alignment}
\end{figure*}

Distributed LLM training requires selecting parallel configurations that fit
device memory and minimize iteration time. Performance depends on coupled
choices in data, tensor, pipeline, context, and expert parallelism
(DP/TP/PP/CP/EP), together with activation recomputation, scheduling, and network
topology~\cite{narayanan2021efficient,grattafiori2024llama3herdmodels}.
Evaluating these choices through full training runs is costly. For example,
DeepSeek-V3 reports 2.788 million H800 GPU-hours for its official training,
corresponding to an estimated US\$5.576 million at US\$2 per
GPU-hour~\cite{deepseekai2025deepseekv3technicalreport}. OpenAI likewise notes
that extensive model-specific tuning is infeasible for training runs at
GPT-4's scale~\cite{openai2024gpt4technicalreport}. Given a model,
training strategy, and hardware platform, performance simulation estimates
runtime and memory requirements to guide configuration selection before
deployment (Figure~\ref{fig:model_resource_alignment}).

Existing simulators are evaluated using different hardware platforms,
software stacks, workloads, and validation targets.
For example, vTrain validates training iteration times on A100
systems using Megatron-DeepSpeed~\cite{bang2024vtrainsimulationframeworkevaluating};
ASTRA-sim2.0 validates its analytical network backend against NCCL
All-Reduce measurements on V100 GPUs~\cite{won2023astrasim};
and SimAI evaluates GPT-3 and LLaMA workloads on A100/H100
clusters with RoCEv2 networks~\cite{wang2025simai}.
These differences make reported prediction errors difficult to
compare directly across studies, motivating shared evaluation
tasks and matched execution conditions.
Evaluating training simulators involves iteration-time accuracy,
internal computation and communication costs, and configuration-selection
quality. These complementary targets call for end-to-end measurements
linked to intermediate profiles from the same training runs.

To address these evaluation challenges, we introduce LM-SimBench,
which brings together a unified simulation framework, a profiling
dataset collected with MindSpeed-LLM~\cite{ascend2024mindspeedllm}
on Ascend NPUs~\cite{liao2021ascend}, and a hierarchical benchmark.
The framework identifies simulator inputs and outputs, the dataset
supplies corresponding measurements, and the benchmark evaluates
prediction accuracy, error diagnosis, and configuration selection.

Our contributions are:
\begin{itemize}
    \item \textbf{A unified simulation framework.} We connect model
    compilation and device binding with execution modeling, computation
    and communication cost estimation, and schedule-level aggregation,
    relating simulator design choices to evaluation targets.
    \item \textbf{A multi-granularity profiling dataset.} We release 348
    feasible configurations across 23 model settings from 10 dense and
    MoE LLM families, covering five parallelism dimensions
    and recomputation. 
    \item \textbf{A hierarchical evaluation benchmark.} System-, component-,
    and strategy-level tasks connect prediction accuracy with error diagnosis
    and configuration selection. Calculon and vTrain demonstrate comparisons of prediction accuracy,
    simulation cost, and diagnostic outputs.
    \item \textbf{Cross-platform protocol validation.} Matched measurements
    for Qwen3-0.6B and Qwen3-1.7B on NVIDIA V100 GPUs and Ascend NPUs
    demonstrate cross-platform evaluation and NPU-guided GPU
    configuration selection.
\end{itemize}

\section{Related Work}
\label{sec:related_work}

\paragraph{Distributed LLM training simulators.}
Existing simulators estimate training time and, in some cases, device-memory
requirements under different parallel
configurations~\cite{IsaevMDV23,qin2026phantora}.
Table~\ref{tab:related_simulator_capabilities} summarizes representative tools
by their operator-time estimation approach, prediction outputs, and
availability of parameter-calibration tools. Analytical estimators apply cost equations with
hardware parameters~\cite{IsaevMDV23,moorethreads2026simumax};
profiling-based estimators use measured operator or kernel
latencies~\cite{duan2024proteus,bang2024vtrainsimulationframeworkevaluating,qin2026phantora};
hybrid estimators combine measured costs with analytical or mechanistic
models~\cite{wang2025simai}. Cost estimation and execution
representation are separate modeling dimensions.
For example, vTrain~\cite{bang2024vtrainsimulationframeworkevaluating}
combines profiled kernel times with analytical inter-node communication
estimates, while SimAI~\cite{wang2025simai} combines kernel profiling
with detailed communication simulation. Lumos~\cite{liang2025lumos}
develops trace-driven performance modeling for large-scale LLM training.

\begin{table*}[t]
\centering
\caption{Representative simulators for distributed LLM training.
Cost estimation refers to operator computation time:
Analytical uses hardware-based cost equations;
Profiling-based directly uses measured operator or kernel latencies;
Hybrid uses analytical models calibrated with operator profiling data.
Memory denotes device-memory usage; Calibration tool indicates
the availability of parameter-fitting tools.}
\label{tab:related_simulator_capabilities}
\small
\setlength{\tabcolsep}{6pt}
\renewcommand{\arraystretch}{1.12}
\begin{tabularx}{\textwidth}{@{}>{\RaggedRight\arraybackslash}Xcccc@{}}
\toprule
\textbf{Simulator} & \textbf{Cost estimation} & \textbf{Time} & \textbf{Memory} & \textbf{Calibration tool} \\
\midrule
Calculon~\cite{IsaevMDV23} & Analytical & $\checkmark$ & $\checkmark$ & $\times$ \\
SimuMax~\cite{moorethreads2026simumax} & Analytical & $\checkmark$ & $\checkmark$ & $\checkmark$ \\
\midrule
Proteus~\cite{duan2024proteus} & Profiling-based & $\checkmark$ & $\checkmark$ & $\times$ \\
vTrain~\cite{bang2024vtrainsimulationframeworkevaluating} & Profiling-based & $\checkmark$ & $\times$ & $\times$ \\
Phantora~\cite{qin2026phantora} & Profiling-based & $\checkmark$ & $\checkmark$ & $\times$ \\
\midrule
ASTRA-sim 2.0~\cite{won2023astrasim,astrasim2026implementation} & Hybrid & $\checkmark$ & $\checkmark$ & $\times$ \\
SimAI~\cite{wang2025simai} & Hybrid & $\checkmark$ & $\times$ & $\times$ \\
Echo~\cite{feng2024echo} & Hybrid & $\checkmark$ & $\times$ & $\checkmark$ \\
\bottomrule
\end{tabularx}
\par\vspace{2pt}
\begin{minipage}{\textwidth}
\footnotesize
ASTRA-sim's memory tracking requires tensor metadata;
Echo's calibration targets its slowdown predictor.
\end{minipage}
\end{table*}

\paragraph{Profiling data and execution traces.}
Existing resources provide measurements and workload descriptions at
different granularities
(Table~\ref{tab:related_artifact_comparison}).
NeuSight~\cite{lee2025neusight} provides operator and end-to-end
latency measurements across Transformer models and GPU platforms.
AcmeTrace~\cite{hu2024characterization} and Alibaba GPU
traces~\cite{alibaba2020clusterdata} record production workloads
and resource usage for cluster-level analysis.
Chakra~\cite{sridharan2023chakra} provides a standardized
execution-trace representation, while AICB~\cite{alibaba2024aicb}
generates parameterized workloads for communication simulation.
LM-SimBench links end-to-end measurements to per-rank computation,
communication, and memory profiles from the same training runs
across controlled parallel configurations, providing a common
measurement basis for simulator calibration and evaluation.

\begin{table*}[t]
\centering
\caption{Related public artifacts and the primary LM-SimBench collection.
The comparison distinguishes measured profiling datasets, cluster traces,
and tools for representing or generating workloads.}
\label{tab:related_artifact_comparison}
\small
\setlength{\tabcolsep}{4pt}
\renewcommand{\arraystretch}{1.12}
\begin{tabularx}{\textwidth}{@{}>{\RaggedRight\arraybackslash}p{0.16\textwidth}>{\RaggedRight\arraybackslash}p{0.17\textwidth}>{\RaggedRight\arraybackslash}X>{\RaggedRight\arraybackslash}p{0.18\textwidth}>{\RaggedRight\arraybackslash}p{0.10\textwidth}@{}}
\toprule
\textbf{Artifact} & \textbf{Type / workloads} & \textbf{Released signals or specifications} & \textbf{Parallelism coverage} & \textbf{Hardware} \\
\midrule
NeuSight~\cite{lee2025neusight}
& Profiling dataset; 5 dense + 1 MoE Transformer
& Operator and end-to-end latency measurements
& DP/TP/PP sweeps
& NVIDIA / AMD GPUs \\
AcmeTrace~\cite{hu2024characterization}
& Cluster trace; production LLM jobs
& Scheduler logs, resource monitoring, timing, and failures
& Job-level observations
& A100 GPUs \\
Alibaba GPU traces~\cite{alibaba2020clusterdata}
& Cluster trace; production GPU jobs
& Job-level resource and timing records
& Job-level observations
& GPUs \\
Chakra~\cite{sridharan2023chakra}
& Execution-trace format
& Graph schema and example traces; collection and replay tools
& Representation schema
& Hardware-agnostic \\
Lumos~\cite{liang2025lumos}
& Modeling toolkit; GPT-3 family
& Kineto and execution-trace tooling
& Modeling methodology
& H100 GPUs \\
AICB~\cite{alibaba2024aicb}
& Workload generator; dense and MoE LLMs
& Parameterized workload specifications
& DP/TP/PP/SP/EP specifications
& GPU-agnostic \\
\midrule
\textbf{LM-SimBench}
& \textbf{Profiling dataset}; 23 settings, 10 dense/MoE families
& \textbf{Six linked views}: iteration time, peak memory, time breakdown,
operator latency, communication events, and matrix kernels
& \textbf{348 configurations}; DP/TP/PP/CP/EP and recomputation
& Ascend NPUs \\
\bottomrule
\end{tabularx}
\end{table*}

\paragraph{Training benchmarks and simulator evaluation.}
MLPerf Training~\cite{mattson2020mlperftraining} standardizes
workloads, quality targets, and reporting for evaluating training
systems. Simulator evaluation addresses a different question:
how accurately predictions reproduce measured system behavior
and how effectively they guide configuration selection.
Existing simulators validate their predictions using different
workloads, platforms, and performance targets, including training
iteration time and collective-communication
latency~\cite{bang2024vtrainsimulationframeworkevaluating,
won2023astrasim,wang2025simai}.
LM-SimBench organizes evaluation into system-, component-, and
strategy-level tasks. These tasks compare overall time and memory
predictions with measurements, assess internal computation and
communication costs, and evaluate the quality of recommended
parallel configurations.

\section{A Unified Framework for Large-Model Training Simulation}
\label{sec:framework}

We describe training simulators through common inputs, outputs, and functional
stages. This framework connects their execution and cost-modeling choices to
measurable targets for overall accuracy, internal cost estimates, and
configuration selection.

\begin{figure*}[t]
    \centering
    \includegraphics[width=\linewidth]{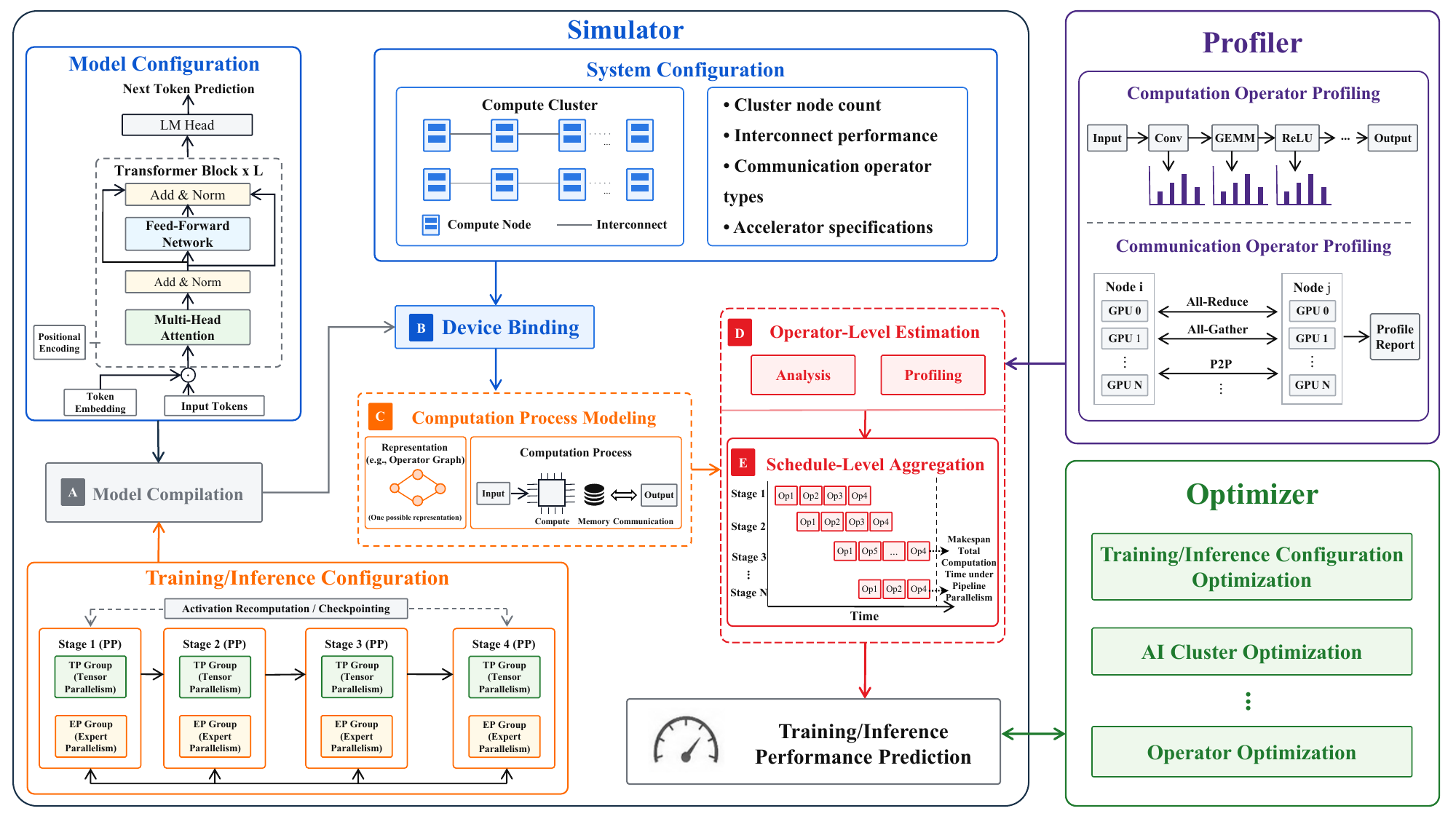}
    \caption{Unified pipeline for large-model training simulation. Model,
    training, and system configurations drive compilation, device binding,
    and computation-process modeling. Computation and communication costs
    from operator-level estimation feed schedule-level aggregation to predict
    iteration time. Profiling supplies measured costs; the optimizer uses
    predictions for configuration search.}
    \label{fig:framework}
\end{figure*}

\subsection{Pipeline and Formal Problem}

A simulation instance is $x=(M,P,N)$: $M$ specifies model architecture and
workload; $P$ specifies parallelism, micro-batch size, recomputation, precision,
fusion, and overlap; and $N$ specifies compute, memory, and interconnect
characteristics. RC and noRC denote configurations with and without activation
recomputation, respectively. The simulator predicts
\begin{equation}
    \mathcal{S}_{\theta}(x)
    =(\widehat{T},\widehat{\mathbf{V}},\widehat{C},\widehat{z}),
    \label{eq:framework-simulator-map}
\end{equation}
where $\theta$ denotes simulator parameters. Outputs comprise iteration time
$\widehat{T}$, per-device peak memory
$\widehat{\mathbf{V}}=(\widehat{V}_d)_{d\in D(N)}$, configuration-selection objective
$\widehat{C}$, and optional diagnostics $\widehat{z}$. $D(N)$ is the device set.

Figure~\ref{fig:framework} defines five functional stages. Stage A compiles
the model under $P$, Stage B assigns devices and parallel groups, and
Stage C constructs the execution representation:
\begin{equation}
    \begin{aligned}
    \bar{M}&=\mathcal{C}(M,P), \\
    \Pi&=\mathcal{B}(\bar{M},P,N), \\
    R&=\mathcal{R}(\bar{M},\Pi,P,N).
    \end{aligned}
    \label{eq:framework-representation}
\end{equation}
Here, $\bar{M}$ is the compiled model and $\Pi$ records placement and
communication groups. $R$ represents execution through analytical phases,
an operator-level directed acyclic graph (DAG), a profiled trace, or a hybrid
schedule.

Stage D estimates the costs of computation and communication units in $R$.
Let $\mathcal{U}(R)=
\mathcal{U}_{\mathrm{op}}(R)\cup\mathcal{U}_{\mathrm{comm}}(R)$ denote these
units. Their predicted latencies are
\begin{equation}
    \widehat{t}_u=
    \begin{cases}
    \mathcal{F}_{\mathrm{op}}(\phi_u,\Pi,N),
        & u\in\mathcal{U}_{\mathrm{op}}(R),\\
    \mathcal{F}_{\mathrm{comm}}(\phi_u,\Pi,N),
        & u\in\mathcal{U}_{\mathrm{comm}}(R),
    \end{cases}
    \label{eq:framework-unit-latency}
\end{equation}
where $\phi_u$ describes tensor shape, FLOPs, transferred bytes, primitive,
and group size. Estimators may be analytical, structure-aware, profiling-based,
or learned. The profiler supplies computation and communication measurements
for cost estimation.

Stage E aggregates unit costs under execution dependencies, resource
constraints, and computation--communication overlap to predict iteration time.
Time and memory predictions are expressed as
\begin{equation}
    \begin{aligned}
    \widehat{T}
    &=\mathcal{A}\!\left(
        R,\Pi,\{\widehat{t}_u\}_{u\in\mathcal{U}(R)},N
      \right),\\
    \widehat{\mathbf{V}}
    &=\mathcal{V}(R,\Pi,N).
    \end{aligned}
    \label{eq:framework-aggregation}
\end{equation}
The aggregation function $\mathcal{A}$ implements phase rules, constrained
graph scheduling, or trace replay, determining which scheduling and overlap
effects are represented. The memory estimator $\mathcal{V}$ uses tensor layouts
and lifetime assumptions in $R$ and $\Pi$, together with $N$.

Table~\ref{tab:main-simulator-map} compares execution representations (Stage C)
and cost estimators (Stage D), linking simulator design to calibration
requirements and intermediate evaluation targets.

\begin{table*}[t]
\centering
\caption{Representative simulators in the unified framework: execution
representations in Stage C and computation and communication cost estimators
in Stage D.}
\label{tab:main-simulator-map}
\small
\setlength{\tabcolsep}{4pt}
\renewcommand{\arraystretch}{1.10}
\begin{tabularx}{\textwidth}{@{}>{\RaggedRight\arraybackslash}p{0.20\textwidth}>{\RaggedRight\arraybackslash}p{0.20\textwidth}>{\RaggedRight\arraybackslash}p{0.27\textwidth}>{\RaggedRight\arraybackslash}X@{}}
\toprule
\textbf{Simulator} & \textbf{Execution} & \textbf{Computation} & \textbf{Communication} \\
\midrule
Calculon~\cite{IsaevMDV23} & Analytical phases & Analytical & Mechanistic \\
SimAI~\cite{wang2025simai} & Workload graph & Profiled kernels / analytical scaling & Mechanistic / packet-level \\
Optimus~\cite{kundu2024performance} & Operator graph & Analytical / structure-aware & Mechanistic \\
ASTRA-sim 2.0~\cite{won2023astrasim} & Workload graph & Profiled / roofline & Mechanistic / packet-level \\
vTrain~\cite{bang2024vtrainsimulationframeworkevaluating} & Execution graph & Profiled CUDA kernels & Profiled / analytical \\
LLMEmu~\cite{yang2025llmemu} & Hybrid & Profiling-based & Trace-driven \\
\bottomrule
\end{tabularx}
\end{table*}

\subsection{Connecting Simulation to Evaluation}

LM-SimBench evaluates overall predictions, internal cost estimates, and
configuration selection using linked measurements from the same training runs:
\begin{equation}
    \mathcal{D}
    =\{(x_i,y_i^{\star},z_i^{\star})\}_{i=1}^{n},
    \label{eq:framework-dataset}
\end{equation}
where $x_i=(M_i,P_i,N_i)$, $y_i^{\star}=(T_i^{\star},\mathbf{V}_i^{\star})$
contains measured time and per-device peak memory, and $z_i^{\star}$ contains
rank-level breakdowns, operator latencies, communication events, and
matrix-kernel records. Configuration and rank identifiers link these views.

System-level tasks
compare time and memory predictions with $y_i^{\star}$. Component-level tasks
compare intermediate estimates with $z_i^{\star}$ at matching units and
aggregation granularity: computation and communication profiles assess Stage D,
while time breakdowns help assess scheduling and overlap in Stage E.
Strategy-level tasks compare candidate configurations under a fixed model
and system, assessing selection quality against measured iteration times
(Section~\ref{sec:benchmark}).

Configuration selection retains strategies that are valid and predicted
to fit on every device:
\begin{equation}
    \begin{aligned}
    I_{\mathrm{feas}}(M,P,N)
    &=\mathbf{1}\{P\in\mathcal{K}(M,N)\}\\
    &\quad\cdot\prod_{d\in D(N)}
    \mathbf{1}\{\widehat{V}_d\leq\mathrm{cap}_d\},
    \end{aligned}
    \label{eq:framework-feasibility}
\end{equation}
where $\mathcal{K}(M,N)$ is the valid strategy space and $\mathrm{cap}_d$ is
device capacity. Scalar peak memory is
$\widehat{V}=\max_{d\in D(N)}\widehat{V}_d$. Ranking retains predicted-feasible
candidates; measured memory supplies the feasibility evaluation labels.

\section{LM-SimBench Dataset}
\label{sec:dataset}

LM-SimBench contains distributed-training profiles for 23 model settings
from 10 dense and MoE LLM families, covering 348 feasible parallel
configurations. A model setting specifies the model and training sequence
length; its configurations vary DP/TP/PP/CP/EP and recomputation settings. The primary
collection uses Ascend 910B3 NPUs with 64\,GiB per device and
MindSpeed-LLM~\cite{ascend2024mindspeedllm}.

Training logs and per-rank Ascend Profiler outputs are processed into six
measurement views (Table~\ref{tab:data_views_tasks}). Shared model-setting,
configuration, and rank identifiers connect end-to-end labels to intermediate
records from the same run. Processing checks configuration names, rank
coverage, units, and schemas. The resulting 26.2\,GB of CSV tables and
550\,GB of raw profiles support both direct evaluation and trace-level analysis.

\begin{table}[t]
\centering
\caption{Six linked measurement views and their evaluation uses.}
\label{tab:data_views_tasks}
\small
\setlength{\tabcolsep}{3pt}
\renewcommand{\arraystretch}{1.10}
\begin{tabularx}{\linewidth}{@{}>{\RaggedRight\arraybackslash}p{0.27\linewidth}>{\RaggedRight\arraybackslash}p{0.24\linewidth}>{\RaggedRight\arraybackslash}X@{}}
\toprule
\textbf{View} & \textbf{Granularity} & \textbf{Evaluation use} \\
\midrule
Iteration time & Configuration & Runtime prediction and ranking \\
Peak memory & Rank & Memory prediction and feasibility \\
Time breakdown & Rank & Bottleneck diagnosis \\
Operator latency & Operator/rank & Computation cost modeling \\
Communication & Event/rank & Communication cost modeling \\
Matrix kernels & Kernel/rank & Matrix-operator prediction \\
\bottomrule
\end{tabularx}
\end{table}

The dataset also includes paired NVIDIA V100 and Ascend 910B3
measurements for Qwen3-0.6B and Qwen3-1.7B under the same schema.
The dataset comprises system telemetry collected from controlled
training runs.

\section{Benchmark Design}
\label{sec:benchmark}

LM-SimBench evaluates system- and component-level predictions,
configuration-selection quality, and simulation cost. Cross-platform tasks
compare paired measurements and evaluate NPU-guided GPU configuration selection.

\subsection{Evaluation Protocol}
\label{subsec:evaluation-protocol}

For each $x=(M,P,N)$, predictions are matched to measured time, memory, and
component costs through configuration, rank, and component identifiers.
Shared-task comparisons use the same configurations and measurement labels,
with sample counts reported for each experiment.

Calibration measurements supply simulator parameters or profiled costs.
Simulator comparisons use disjoint calibration and test configurations;
calibration-coverage experiments vary the calibration set while fixing the
test set. Platform-calibration experiments fit parameters to each stack's
measurements and report the resulting calibration errors.

\subsection{System- and Component-Level Prediction}
\label{subsec:system-level-tasks}
\label{subsec:component-level-tasks}

\paragraph{Time and memory.}
Iteration-time predictions are compared with configuration-level measurements.
Peak-memory error is computed per rank for platform calibration and on the
maximum across ranks for configuration-level diagnostics. Configuration maxima
also determine memory rankings and capacity decisions. Feasibility evaluation
reports agreement with measured memory and retention of measured-feasible
configurations.

\paragraph{Component costs.}
Computation targets include matrix-kernel latency per call, stage computation
time, and recomputation-stage time. Predictions are paired with measurements
at the corresponding operator or stage granularity. Communication targets
include event durations and non-overlapped communication time from time breakdowns. Aggregated estimates are compared at matching configuration and
rank granularity, with durations expressed in common units.

Prediction metrics are mean absolute error (MAE), symmetric mean absolute
percentage error (sMAPE), and $R^2$. For matched predictions $\widehat{y}_i$
and measurements $y_i$, sMAPE averages
$200|\widehat{y}_i-y_i|/(|\widehat{y}_i|+|y_i|+\epsilon)$,
with $\epsilon=10^{-12}$. Platform calibration reports mean absolute
percentage error (MAPE) for iteration time and per-rank memory, with memory
ranking evaluated on configuration maxima.

\subsection{Configuration Selection and Simulation Cost}
\label{subsec:strategy-ranking}

For a fixed model and system, configuration selection applies the capacity
criterion in Eq.~\ref{eq:framework-feasibility} and ranks candidates by
predicted iteration time.
Spearman $\rho$ and Kendall $\tau$ compare predicted and measured orderings.
Top-$k$ hit records whether the measured optimum enters the recommended set;
Top-$k$ set overlap is the fraction shared with the measured top-$k$ set.

Let $\mathcal{C}$ be the evaluation candidate set, $\mathcal{S}_k$ the
recommended $k$ configurations, and $T^{\star}(c)$ the measured iteration time.
Shortlist regret is
\begin{equation}
    \operatorname{Regret}@k
    =100\left(
    \frac{\min_{c\in\mathcal{S}_k}T^{\star}(c)}
         {\min_{c\in\mathcal{C}}T^{\star}(c)}-1\right)\%,
    \label{eq:benchmark-shortlist-regret}
\end{equation}
which evaluates the fastest measured member of the shortlist. At $k=1$,
it gives the selected configuration's time penalty. Simulation cost is evaluated using the median simulation time per
configuration, measured as software wall-clock time in the same CPU and
software environment.

\subsection{Cross-Platform Evaluation}
\label{subsec:cross-platform-metrics}

Paired comparisons match model settings and configuration identifiers across
GPU and NPU stacks and report the matched sample count. Time and memory
rankings are compared using Spearman $\rho$, Kendall $\tau_b$, Top-1 agreement,
and Top-$k$ set overlap, both pooled and stratified by RC/noRC.
Matched RC/noRC pairs assess the direction of time and memory changes.

Matrix problems are matched by batch size, dimensions $(m,n,k)$, and recorded
input types. Per-call latencies are pooled across paired configurations and
ranks to obtain a median per problem and stack. These medians define latency
rankings, compared by Spearman $\rho$ and slowest-problem Top-$k$ overlap.
Operator-category and communication-primitive durations are summed across ranks per configuration
and normalized into time-share vectors. The vectors are aligned over the
union of categories, with zero entries for absent categories. Spearman
correlation and Jensen--Shannon divergence (JSD) are computed per configuration
and summarized by their medians.

NPU-guided GPU selection ranks the paired candidates by measured NPU time
and retains the fastest $k$. GPU measurements evaluate optimum inclusion
and shortlist regret (Eq.~\ref{eq:benchmark-shortlist-regret}), using the
GPU-fastest configuration in the paired candidate set as the reference.

\section{Results}
\label{sec:results}

We evaluate simulator accuracy, configuration selection, and simulation
cost, examine computation, memory, and communication through intermediate
profiles, and assess platform calibration and NPU-guided GPU configuration
selection.

\subsection{Simulator Accuracy, Selection, and Cost}
\label{subsec:simulator-comparison}

\paragraph{Setup.}
We evaluate Calculon and vTrain on Qwen1.5-1.8B at 8K tokens with
CP${=}1$ on the NPU stack. Calculon uses analytical cost models, while
vTrain combines an execution graph with LM-SimBench profiles supplied
through an adapter. Ten configurations calibrate vTrain, and ten disjoint
configurations form the test set. We compare iteration-time prediction,
configuration ranking, and simulation cost on the six test configurations
for which both simulators produce valid predictions.

\begin{table}[t]
\centering
\caption{Calculon and vTrain on six common test configurations of
Qwen1.5-1.8B (8K tokens). Simulation time is median wall-clock time per configuration.}
\label{tab:main-simulators}
\small
\setlength{\tabcolsep}{3pt}
\renewcommand{\arraystretch}{1.10}
\begin{tabularx}{\linewidth}{@{}>{\RaggedRight\arraybackslash}Xrr@{}}
\toprule
\textbf{Metric} & \textbf{Calculon} & \textbf{vTrain} \\
\midrule
Iteration-time MAE (s) $\downarrow$ & 1.059 & \textbf{0.207} \\
Iteration-time sMAPE (\%) $\downarrow$ & 8.88 & \textbf{1.86} \\
Iteration-time $R^2$ $\uparrow$ & 0.630 & \textbf{0.990} \\
Ranking Spearman $\rho$ $\uparrow$ & 0.943 & \textbf{1.000} \\
Ranking Kendall $\tau$ $\uparrow$ & 0.867 & \textbf{1.000} \\
Top-1 hit / regret & 1 / 0\% & 1 / 0\% \\
Top-3 set overlap & 3/3 & 3/3 \\
Simulation time (ms) $\downarrow$ & \textbf{1.78} & 25.46 \\
\bottomrule
\end{tabularx}
\end{table}

\begin{table}[t]
\centering
\caption{Calculon on platform calibration configurations. Entries report
MAPE (\%) / Spearman $\rho$. Memory error uses per-rank peaks; ranking
uses the maximum peak across ranks.}
\label{tab:main-calibration}
\small
\setlength{\tabcolsep}{3pt}
\renewcommand{\arraystretch}{1.08}
\begin{tabularx}{\linewidth}{@{}Xrr@{}}
\toprule
\textbf{Qwen3 / stack} & \textbf{Time} & \textbf{Memory} \\
\midrule
0.6B / V100 & 10.28 / 0.861 & 15.99 / 0.993 \\
0.6B / 910B3 & 10.19 / 0.950 & 13.20 / 0.986 \\
1.7B / V100 & 11.22 / 0.853 & 12.56 / 0.993 \\
1.7B / 910B3 & 7.97 / 0.935 & 12.55 / 0.994 \\
\bottomrule
\end{tabularx}
\end{table}

\begin{table*}[t]
\centering
\caption{Cross-platform transfer from Ascend 910B3 to NVIDIA V100.
Shortlist regret uses GPU time.
Correlations compare measured profiles on the two stacks.}
\label{tab:main-transfer}
\small
\setlength{\tabcolsep}{5pt}
\renewcommand{\arraystretch}{1.08}
\begin{tabularx}{\textwidth}{@{}l>{\RaggedRight\arraybackslash}Xcc@{}}
\toprule
\textbf{Level} & \textbf{Metric} & \textbf{Qwen3-0.6B} & \textbf{Qwen3-1.7B} \\
\midrule
Coverage & Paired complete configurations & 15 & 11 \\
Strategy & Iteration-time rank $\rho$ & 0.718 & 0.664 \\
         & NPU Top-1 selection: GPU regret $\downarrow$ & 0.0\% & 6.9\% \\
         & GPU optimum in NPU Top-3 & \textbf{Yes} & \textbf{Yes} \\
         & NPU Top-3 shortlist: GPU regret $\downarrow$ & \textbf{0.0\%} & \textbf{0.0\%} \\
\midrule
Memory & Peak-memory rank $\rho$ & \textbf{0.982} & \textbf{0.991} \\
       & Lowest-memory Top-3 overlap & 3/3 & 3/3 \\
       & RC time/memory direction agreement & 6/6 pairs & 3/3 pairs \\
\midrule
Component & Matrix median-latency rank $\rho$ & \textbf{0.983} & \textbf{0.987} \\
          & Slowest matrix-problem Top-10 overlap & 8/10 & 10/10 \\
          & Communication-primitive median $\rho$ & 0.949 & 0.872 \\
          & Operator-category median $\rho$ & 0.503 & 0.531 \\
          & Non-overlapped communication share, GPU/NPU & 17.0\% / 28.5\% & 10.9\% / 30.0\% \\
\bottomrule
\end{tabularx}
\end{table*}

\paragraph{Accuracy, selection, and cost.}
The two simulators exhibit distinct accuracy and execution-cost
characteristics (Table~\ref{tab:main-simulators}). vTrain achieves
1.86\% iteration-time sMAPE and Spearman $\rho=1.000$, compared with
8.88\% and 0.943 for Calculon. Both identify the measured fastest
configuration and recover the measured top-three set. Calculon's median
simulation time is 1.78\,ms per configuration, compared with 25.46\,ms
for vTrain, a $14.3\times$ speedup.

\paragraph{Calibration coverage.}
Broader configuration coverage improves vTrain's prediction accuracy.
On a fixed set of eight test configurations, calibration with ten
TP/DP/PP profiles yields 2.08\% sMAPE, compared with 11.77\% using
six TP-endpoint profiles. The 9.69-percentage-point reduction demonstrates
the value of sampling multiple parallelism dimensions for profiling-based
simulation.

\begin{figure*}[t]
\centering
\includegraphics[width=\textwidth]{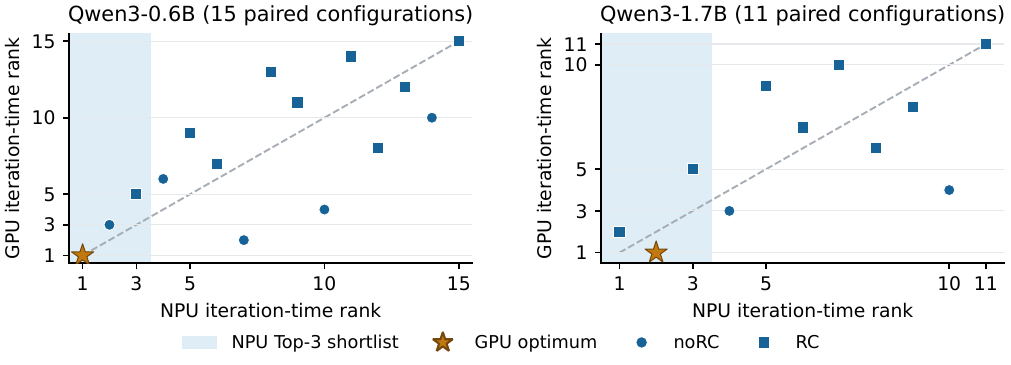}
\caption{NPU profiles guide GPU configuration selection. Each point is a
paired configuration ranked by measured iteration time (rank 1 is fastest).
The shaded NPU Top-3 shortlist contains the GPU optimum (star) for both
models. The dashed line denotes identical ranks.}
\label{fig:cross-platform-shortlist}
\end{figure*}

\subsection{Diagnosis with Intermediate Profiles}
\label{subsec:diagnostic-results}

\paragraph{Computation.}
Operator- and stage-level profiles characterize computation predictions
at their respective granularities. Calculon's matrix-operator predictions
achieve $R^2=0.817$, with 0.109\,ms MAE and 43.34\% sMAPE. For vTrain,
stage-computation sMAPE is 11.60\%, and recomputation-stage sMAPE is 2.75\%.

\paragraph{Memory.}
Calculon's memory model retains 13 of the 20 measured-feasible
configurations, with a peak-memory MAE of 6.92\,GiB on the retained set.
Paired predictions and measurements reveal overestimates of
13--29\,GiB in several noRC configurations, linking memory accounting
to candidate retention.

\paragraph{Communication.}
Time breakdowns connect communication estimates with the overlap
observed during training. vTrain's non-overlapped communication prediction
has 136.20\% sMAPE. The associated communication profiles expose the
difference between serialized communication estimates and observed overlap,
providing a concrete target for refining communication scheduling.

\subsection{Simulator Calibration across Platforms}
\label{subsec:platform-calibration}

\paragraph{Setup.}
We collect paired profiles for Qwen3-0.6B and Qwen3-1.7B at 4K tokens on
8$\times$NVIDIA V100 GPUs (32\,GiB, Megatron-LM) and
8$\times$Ascend 910B3 NPUs (64\,GiB, MindSpeed-LLM). Each model uses 18
DP/TP/PP/recomputation candidates with CP${=}1$, micro/global batch sizes
1/32, and a distributed optimizer. We profile the sixth iteration after
five warm-up steps. V100 uses FP16; 910B3 uses BF16, fused kernels, and
sequence parallelism. The paired evaluation contains 15 complete
configurations for Qwen3-0.6B and 11 for Qwen3-1.7B.

\paragraph{Platform calibration.}
We apply a common Calculon implementation incorporating Qwen3 architecture,
pipeline scheduling, kernel-launch latency, and rank-level memory
accounting. For each platform, a single parameter set for compute
efficiency, launch latency, communication, and runtime memory is calibrated
jointly across both models and their candidate configurations.

On the calibration configurations, iteration-time MAPE ranges from
7.97\% to 11.22\%, with ranking $\rho=0.853$--$0.950$; peak-memory
ranking reaches $\rho=0.986$--$0.994$
(Table~\ref{tab:main-calibration}). Within each of the four
model$\times$RC/noRC groups, GPU measurements, NPU measurements, and
both calibrated simulators identify the same Top-1 configuration.

\subsection{Cross-Platform Configuration Selection}
\label{subsec:cross-platform-results}

\paragraph{NPU-guided GPU selection.}
Ranking the paired configurations by measured NPU iteration time produces
a three-candidate shortlist containing the GPU-fastest configuration for
both models (Table~\ref{tab:main-transfer};
Figure~\ref{fig:cross-platform-shortlist}). This reduces the candidate sets
from 15 to 3 and from 11 to 3. Selecting the fastest shortlisted
configuration using GPU measurements yields zero Top-3 shortlist regret
in both cases. Direct selection of the NPU-fastest configuration gives
GPU regret of 0.0\% and 6.9\%, respectively.

\paragraph{Memory and recomputation trends.}
Cross-platform peak-memory rankings achieve $\rho=0.982$ and $0.991$,
with identical three lowest-memory configurations. Across all nine matched
RC/noRC pairs, recomputation increases iteration time and reduces peak
memory on both platforms. These consistent trends support memory-oriented
candidate screening across the two stacks.

\paragraph{Component-level execution characteristics.}
Matrix-problem latency rankings agree across platforms, with
$\rho=0.983$ and $0.987$ and slowest-problem Top-10 overlaps of 8/10 and
10/10. Communication-primitive time shares have median rank correlations
of 0.949 and 0.872. Operator-category correlations are 0.503 and 0.531,
and non-overlapped communication accounts for 17.0\%/10.9\% of runtime
on V100 and 28.5\%/30.0\% on 910B3. The profiles identify expensive
shared matrix problems and characterize each stack's operator composition
and communication overlap.

\section{Conclusion}
\label{sec:limitations}

LM-SimBench connects a unified simulation framework, a multi-granularity
profiling dataset, and a hierarchical evaluation benchmark for distributed
LLM training. Its linked measurements support simulator calibration,
prediction assessment, and diagnosis of computation, memory, and
communication costs. Evaluations of Calculon and vTrain demonstrate how
the same benchmark captures prediction accuracy, configuration-selection
utility, and simulation cost. GPU--NPU experiments show that NPU profiles
preserve useful memory and matrix-operator rankings and identify shortlists
containing the GPU-optimal configurations. Extending workload
and accelerator coverage will broaden this common basis for training-system
simulation and optimization.

\clearpage
\section*{LLM Usage Statement}

We used generative AI tools to assist with the writing and presentation of
this manuscript: improving its organization, clarity, and academic English, and
checking internal consistency and layout. Generative AI tools were not used to
formulate the research hypotheses, to design the methodology or the
experiments, to implement the released code or the evaluated simulators, to
generate or process the measured profiling data, or to interpret the results.
All AI-assisted text was reviewed by the authors, who verified every technical
claim and reported value and take responsibility for the final content of this
work.

{\small
\bibliographystyle{ieeenat_fullname}
\bibliography{references}
}
\clearpage
\twocolumn[{\centering\Large\bfseries Appendices\par\vspace{1em}}]
\raggedbottom
\makeatletter
\setlength{\@fptop}{0pt}
\setlength{\@dblfptop}{0pt}
\setlength{\@fpsep}{14pt}
\setlength{\@dblfpsep}{14pt}
\makeatother
\setcounter{topnumber}{3}
\setcounter{dbltopnumber}{3}
\renewcommand{\topfraction}{0.92}
\renewcommand{\dbltopfraction}{0.92}
\renewcommand{\textfraction}{0.06}
\renewcommand{\floatpagefraction}{0.70}
\renewcommand{\dblfloatpagefraction}{0.70}
\appendix
\section*{Overview and Resources}
\label{app:resources}

These appendices follow the main paper from framework and dataset design
to simulator evaluation and cross-platform application.
Appendix~\ref{app:related_work} places LM-SimBench among profiling resources
and performance-modeling tools. Appendix~\ref{app:unified-simulation-framework}
explains the simulation pipeline and modeling choices, and
Appendix~\ref{app:dataset_details} documents the collection, processing,
and file schema of the six linked measurement views.

Appendix~\ref{app:cross_platform_vtrain} reports simulator accuracy,
configuration selection, simulation cost, and calibration coverage.
Appendix~\ref{app:representative_profiling_cases} uses single-node and
multi-node cases to connect predictions with rank-, communication-, and
operator-level measurements.
Appendix~\ref{app:cross_platform} evaluates platform-specific calibration
and NPU-guided GPU configuration selection using paired measurements.

\paragraph{Data and code.}
The resources comprise:
\begin{itemize}
    \setlength{\itemsep}{2pt}
    \setlength{\parskip}{0pt}
    \item \textbf{Processed dataset:} the six linked measurement views,
    manifests, and field documentation.
    \item \textbf{Raw profiles:} the original profiling traces for
    detailed execution analysis.
    \item \textbf{Code:} benchmark evaluation, preprocessing scripts,
    and usage examples.
\end{itemize}
Dataset organization, documentation, and licenses are described in
Appendix~\ref{app:dataset_details}.

\section{Additional Related Work}
\label{app:related_work}

This section expands Section~\ref{sec:related_work} by relating profiling
resources, performance predictors, and training simulators to the evaluation
tasks in LM-SimBench. The organizing distinction is between the measurements
an artifact provides and the modeling choices a simulator uses: linked
measurements allow those choices to be assessed at both the output and
component levels.

\subsection{Profiling Datasets and Trace Releases}
\label{app:profiling_dataset_comparison}

Table~\ref{tab:related_artifact_comparison} compares workload coverage,
released signals, parallelism dimensions, and hardware across related
artifacts based on their released content. The LM-SimBench entry covers the
primary NPU collection; Appendix~\ref{app:cross_platform} reports the matched
GPU supplement.

\textbf{Performance-prediction datasets.}
NeuSight~\cite{lee2025neusight} is the closest artifact in purpose: it releases
operator-level and end-to-end latencies for six Transformer models across
seven NVIDIA and AMD GPUs to support training-performance prediction.
LM-SimBench provides per-rank memory and communication-event measurements,
with a configuration sweep covering context and expert
parallelism and activation recomputation across dense and MoE models.

\textbf{Cluster operational traces.}
AcmeTrace~\cite{hu2024characterization} and the Alibaba GPU cluster
traces~\cite{alibaba2020clusterdata} expose production-job behavior through
scheduler logs, resource monitoring, timing, and failure records. Their
job-level observations support scheduling and cluster-management studies.
LM-SimBench instead associates individual training configurations with
rank-level and operator-level execution measurements.

\textbf{Trace formats and modeling toolkits.}
Chakra~\cite{sridharan2023chakra} standardizes graph-based execution traces
and provides collection, generation, and replay tools.
Lumos~\cite{liang2025lumos} develops trace-driven performance modeling, while
AICB~\cite{alibaba2024aicb}, the workload generator accompanying SimAI,
provides parameterized LLM workload specifications. These tools address
representation and modeling; LM-SimBench supplies measured profiles and
labels for evaluating predictions at multiple granularities.

\subsection{Neural-Network Performance Prediction}

Analytical predictors estimate execution costs from model structure,
operation counts, and hardware characteristics. Roofline relates arithmetic
intensity to compute and memory limits~\cite{williams2009roofline};
SCALE-Sim and Accel-Sim model systolic-array and GPU
execution~\cite{samajdar2019scalesimsystoliccnnaccelerator,khairy2020accelsim}.
Profiling-based approaches measure representative operators and combine
their costs into workload-level estimates. Paleo combines analytical models
with calibrated measurements~\cite{qi2017paleo}, and NeuralPower predicts
CNN-layer runtime, power, and energy~\cite{cai2017neuralpowerpredictdeployenergyefficient}.
Subsequent predictors extend these ideas across accelerators and distributed
settings~\cite{You_2017,samajdar2020systematic,yu2021habitat}.
LM-SimBench provides distributed LLM training measurements for evaluating
both aggregate predictions and their intermediate operator costs.

\subsection{Large-Model System Simulators}

Training simulators combine analytical, graph-based, and measurement-based
modeling~\cite{GuoTZYJWZL24,svedas2025survey}.
Calculon estimates iteration time and memory under Megatron-style
parallelism~\cite{IsaevMDV23,shoeybi2019megatron}, while SimAI connects
framework and network modeling for large-scale LLM
training~\cite{wang2025simai}. Graph- and task-based systems represent
dependencies and scheduling explicitly: ASTRA-sim emphasizes distributed
execution, collective communication, and network
topology~\cite{Won_2023,won2023astrasim}, and Optimus uses operator graphs
for workload and accelerator analysis~\cite{kundu2024performance}.

Measurement-based systems use profiles to estimate execution costs.
vTrain combines an execution graph with profiled kernel times and
profiled or analytical communication estimates~\cite{bang2024vtrainsimulationframeworkevaluating}; LLMEmu uses
GPU-kernel profiles to emulate distributed training on CPU
clusters~\cite{yang2025llmemu}. Habitat and ScaleDL learn performance
relationships across hardware or distributed
settings~\cite{yu2021habitat,wang2025scaledl}. Related serving tools,
including LLMCompass, LLMServingSim, and InferSim, address deployment
performance and request-level behavior~\cite{zhang2023hardwareevaluationframeworklarge,
Cho_2024,group2025infersim}.

LM-SimBench provides common measurements and evaluation tasks for these
different modeling choices. Appendix~\ref{app:unified-simulation-framework}
maps representative training simulators by execution representation,
computation estimator, and communication model.

\subsection{System Benchmarks and Configuration Search}

MLPerf Training and Inference standardize workloads, quality targets, and
reporting procedures for system-level
comparisons~\cite{mattson2020mlperftraining,reddi2020mlperfinference}.
LM-SimBench complements this role by providing reusable training profiles
for time and memory prediction, component-level diagnostics, and
configuration ranking.

Configuration search turns performance estimates into training-system
decisions. General search methods use random sampling, multi-fidelity
allocation, and Bayesian optimization to allocate evaluation
budgets~\cite{Yang_2020,bischl2021hyperparameteroptimizationfoundationsalgorithms,
JMLR:v13:bergstra12a,li2018hyperbandnovelbanditbasedapproach,
falkner2018bohbrobustefficienthyperparameter}. In distributed training,
parallelism, recomputation, scheduling, and placement determine runtime and
memory requirements. LM-SimBench evaluates this use of simulation through
configuration rankings, optimum inclusion, and selection regret, connecting
prediction quality to the configurations a user would choose.

\FloatBarrier
\section{Unified Simulation Framework}
\label{app:unified-simulation-framework}

This section expands the five-stage pipeline in
Section~\ref{sec:framework}. It follows a simulation instance from its input
configurations to execution representation, unit-cost estimates, and
aggregated predictions, then relates these choices to representative
simulators and configuration selection.

\subsection{Simulation Inputs and Execution Pipeline}

\textbf{Simulation inputs.}
We describe the target scenario with three inputs. The model configuration
$M$ records architectural and workload parameters such as layer types, hidden
dimensions, sequence length, global batch size, and parameter count. The system
configuration $N$ specifies the target hardware, including compute capability,
memory capacity, inter-device bandwidth, and topology. The training
configuration $P$ specifies data, tensor, pipeline, context, and expert
parallelism; micro-batch size; recomputation; offloading; numerical precision; operator fusion;
and computation--communication overlap.

\textbf{Performance prediction problem.}
Given $(M,P,N)$, the simulator $\mathcal{S}_\theta$ predicts the performance
of one training iteration. Following Eq.~\ref{eq:framework-simulator-map}, its
outputs are iteration time $\widehat{T}$, per-device peak memory
$\widehat{\mathbf{V}}=(\widehat{V}_d)_{d\in D(N)}$, a configuration-selection objective
$\widehat{C}$, and optional diagnostics $\widehat{z}$. Below, $\widehat{T}$ is
the representative metric, and
$\widehat{V}=\max_{d\in D(N)}\widehat{V}_d$ denotes scalar peak memory when
required.

\textbf{Stages A--C: constructing the execution representation.}
For a training configuration $P$, model compilation transforms the abstract
model into
\begin{equation}
    \bar{M} = \mathcal{C}(M,P),
    \label{eq:app-model-compilation}
\end{equation}
where $\mathcal{C}(\cdot)$ partitions or replicates components according to the
selected parallel strategy. Tensor parallelism partitions parameters within
operators, pipeline parallelism assigns blocks to stages, and data parallelism
replicates the model across groups. Device binding then assigns each component
of $\bar{M}$ to devices and communication groups:
\begin{equation}
    \Pi = \mathcal{B}(\bar{M},P,N),
    \label{eq:app-device-binding}
\end{equation}
where $\Pi$ records placement and parallel-group membership, allowing the
simulator to distinguish intra-node from inter-node communication.

The bound model is then transformed into an execution representation:
\begin{equation}
    R = \mathcal{R}(\bar{M},\Pi,P,N),
    \label{eq:app-execution-representation}
\end{equation}
where $R$ may be an analytical phase decomposition, an operator-level
dependency graph, a profiled trace, or a hybrid schedule. It captures the
execution order, synchronization constraints, and computation--communication
structure induced by pipeline scheduling, tensor partitioning, recomputation,
and overlap.

\textbf{Stages D--E: estimating and aggregating costs.}
Stage D estimates the latency of each computation or communication unit in
$\mathcal{U}(R)=\mathcal{U}_{\mathrm{op}}(R)\cup\mathcal{U}_{\mathrm{comm}}(R)$.
Stage E aggregates these estimates under the ordering and resource constraints
represented by $R$ and $\Pi$. As in Eq.~\ref{eq:framework-unit-latency}, the
predicted latency of unit $u$ is
\begin{equation}
    \widehat{t}_u=
    \begin{cases}
    \mathcal{F}_{\mathrm{op}}(\phi_u,\Pi,N), & u\in\mathcal{U}_{\mathrm{op}}(R),\\
    \mathcal{F}_{\mathrm{comm}}(\phi_u,\Pi,N), & u\in\mathcal{U}_{\mathrm{comm}}(R),
    \end{cases}
    \label{eq:app-operator-estimation}
\end{equation}
where $\phi_u$ records features such as tensor shape, FLOPs, bytes transferred,
primitive type, group size, and placement; $\Pi$ provides placement and group
membership; and $N$ provides the required hardware parameters. The estimator
may be analytical, structure-aware, profiling-based, or learned. This
cost-estimation choice is separate from the execution representation: for
example, an operator graph can use either analytical or measured latencies.

Stage E combines these unit costs with dependencies, resource constraints,
and overlap to obtain iteration time:
\begin{equation}
    \widehat{T} =
    \mathcal{A}\!\left(R,\Pi,\{\widehat{t}_u\}_{u\in\mathcal{U}(R)},N\right).
    \label{eq:app-network-estimation}
\end{equation}
Peak memory is estimated from the execution representation and device
placement, as in Eq.~\ref{eq:framework-aggregation}:
\begin{equation}
    \widehat{\mathbf{V}}=\mathcal{V}(R,\Pi,N),
    \label{eq:app-memory-aggregation}
\end{equation}
Here, $\mathcal{A}(\cdot)$ aggregates time over the execution representation,
and $\mathcal{V}(\cdot)$ estimates peak memory from the tensor-layout and
lifetime assumptions in $R$ and $\Pi$, together with system characteristics
in $N$. The indicator $I_{\mathrm{feas}}(M,P,N)$ from
Eq.~\ref{eq:framework-feasibility} checks strategy validity and predicted
memory against device capacity. Measured memory provides the reference for
evaluating this predicted feasibility.

\textbf{Memory components.}
The principal tensor-memory components are parameters, activations,
gradients, and optimizer states. Their contribution on device $d$ is
\begin{equation}
    \widehat{V}_{\mathrm{tensor},d} = \widehat{V}_{\mathrm{param},d} + \widehat{V}_{\mathrm{act},d} + \widehat{V}_{\mathrm{grad},d} + \widehat{V}_{\mathrm{opt},d},
    \label{eq:app-memory-footprint}
\end{equation}
where $\widehat{V}_{\mathrm{param},d}$, $\widehat{V}_{\mathrm{act},d}$, $\widehat{V}_{\mathrm{grad},d}$, and $\widehat{V}_{\mathrm{opt},d}$ denote the memory used by parameters, activations, gradients, and optimizer states on device $d$, respectively.
Their sizes and lifetimes depend on architecture, sequence length, batch
size, precision, parallelism, and recomputation or offloading. The memory
estimator combines these terms with its runtime-memory accounting to obtain
$\widehat{V}_d$; rank-level measurements provide the corresponding evaluation
targets.

\begin{table*}[t]
\centering
\caption{Representative training simulators characterized by execution
representation (Stage C), computation and communication estimators (Stage D),
artifact availability, and modeling focus.}
\label{tab:simulator_mapping}
\small
\setlength{\tabcolsep}{4.5pt}
\renewcommand{\arraystretch}{1.12}
\begin{tabularx}{\linewidth}{>{\RaggedRight\arraybackslash}p{0.25\linewidth}cccc>{\RaggedRight\arraybackslash}X}
\toprule
\textbf{Simulator} & \textbf{Workflow} & \textbf{Comp.} & \textbf{Comm.} & \shortstack{\textbf{Public}\\\textbf{artifact}} & \textbf{Primary focus} \\
\midrule
Calculon~\cite{IsaevMDV23}        & A   & An      & Mech       & \checkmark & Analytical iteration-time and memory estimation for LLM training. \\
SimAI~\cite{wang2025simai}        & G   & Pr/An   & Mech/Pkt       & \checkmark & Full-stack large-scale training simulation across framework and network layers. \\
Optimus~\cite{kundu2024performance} & G   & An/St   & Mech       & $\times$   & Operator-graph modeling for workload and accelerator microarchitecture exploration. \\
ASTRA-sim 1.0~\cite{Won_2023}     & H   & An      & Mech/Pkt   & \checkmark & Network-topology and collective-communication simulation. \\
ASTRA-sim 2.0~\cite{won2023astrasim} & G & Pr/An   & Mech/Pkt   & \checkmark & Detailed workload graphs combined with network-level simulation. \\
vTrain~\cite{bang2024vtrainsimulationframeworkevaluating} & G   & Pr      & Pr/An      & \checkmark & Graph-based training simulation calibrated by real-system measurements. \\
LLMEmu~\cite{yang2025llmemu}      & H   & Pr      & Trace      & \checkmark & Kernel-profiled GPU emulation of distributed LLM training on CPU clusters. \\
Habitat~\cite{yu2021habitat}      & A   & Pr/Data & Data       & \checkmark & Machine learning (ML)-based runtime prediction across unseen GPU generations. \\
ScaleDL~\cite{wang2025scaledl}    & A   & Data    & Data       & $\times$   & Data-driven performance prediction for distributed training. \\
Paleo~\cite{qi2017paleo}          & A   & An      & Mech       & \checkmark & Early CNN training-time estimator for GPU clusters. \\
\bottomrule
\end{tabularx}
\par\smallskip
{\footnotesize\raggedright
G: graph-based; A/An: analytical; H: hybrid; St: structure-aware;
Pr: profiling-based; Data: data-driven; Mech: white-box mechanistic;
Trace: trace-driven; Pkt: packet-level.\par}
\end{table*}

\subsection{Modeling Choices and Simulator Mapping}

The following choices determine the execution detail and cost of a
simulator. Table~\ref{tab:simulator_mapping} maps representative systems by
execution representation (Stage C) and computation and communication
estimators (Stage D), together with artifact availability and modeling focus.

\textbf{Large-model abstraction.}
For the decoder-only Transformers in LM-SimBench, $M$ records the layer
count, hidden and feedforward dimensions, attention heads, sequence length,
and batch size. These parameters determine the sizes of computation and
memory operations; MoE models additionally specify routing and sparsity.
Appendix~\ref{app:configuration_coverage} lists the profiled workloads.

\begin{figure}[h]
    \centering
    \includegraphics[width=0.85\linewidth]{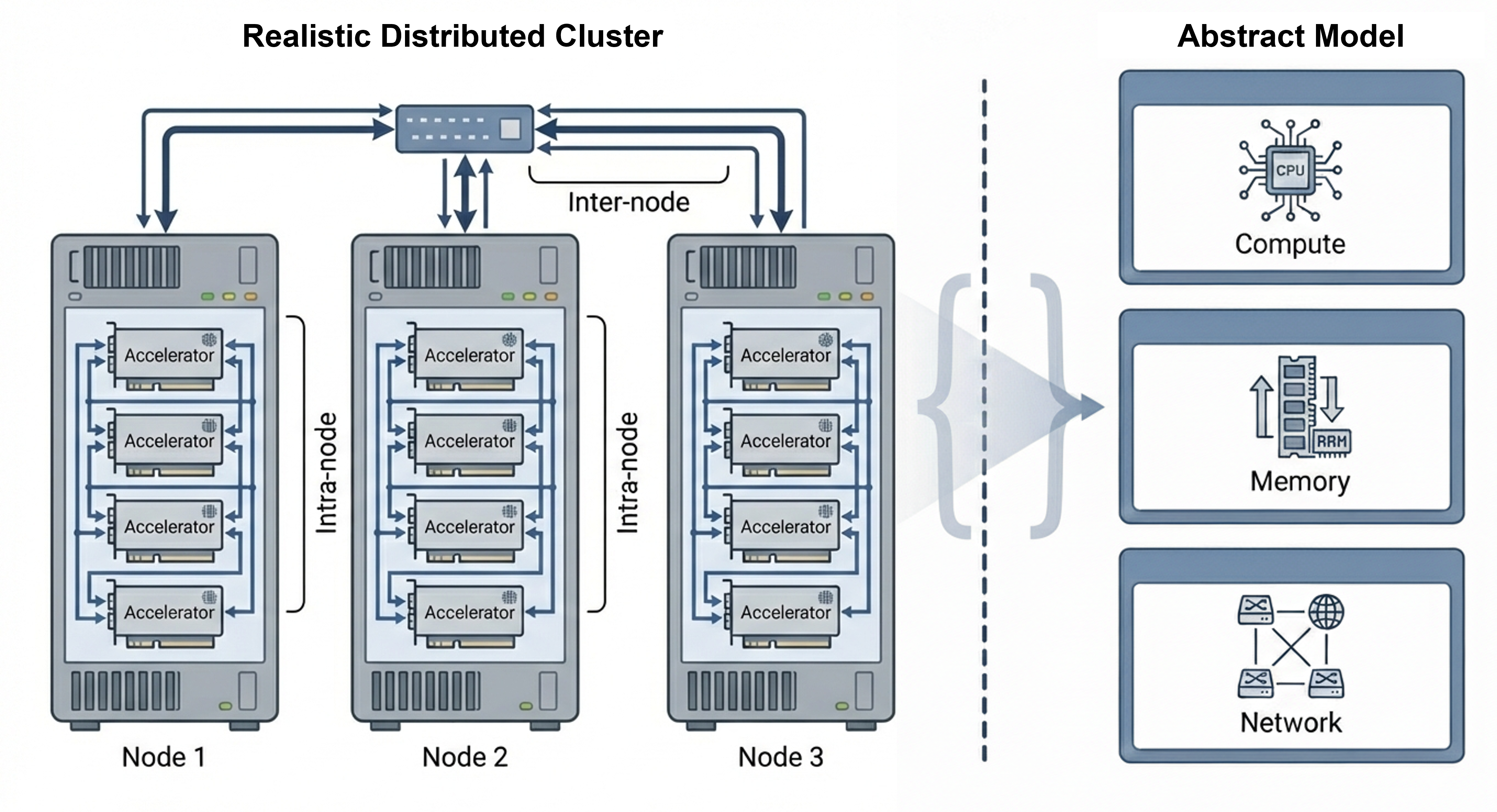}
    \caption{Simplified cluster abstraction. The physical cluster is represented by compute, memory, and network components.}
    \label{fig:app-cluster-abstraction}
\end{figure}

\textbf{Cluster and hardware abstraction.}
Cluster modeling reduces the physical environment to three
performance-relevant components: compute, memory, and network
(Figure~\ref{fig:app-cluster-abstraction}).

Compute is characterized by data type, peak FLOPs, and empirical efficiency;
memory by capacity, bandwidth, and efficiency; and the network by bandwidth,
latency, topology, and communication efficiency. The simulator combines peak
capabilities with efficiency factors to obtain effective performance.

Measured operator and communication latencies can calibrate or replace
analytical cost estimates for the target platform.

\textbf{Training workflow modeling.}
Workflow modeling specifies the temporal order of computation and
communication after compilation and device binding. In an operator-level
graph, nodes represent computation or communication and edges encode data
dependencies. A training graph typically includes the forward pass, backward
pass, gradient synchronization, and parameter update.

Workflow granularity determines which execution effects can be represented.
Graph-based models construct an operator-level DAG for fine-grained timing and
dependency analysis. Analytical models reduce execution to phases such as
forward, backward, communication, and optimizer update; they are faster but
may miss overlap and synchronization. Hybrid models represent critical or
communication-heavy segments explicitly and estimate the remainder
analytically.

\textbf{Computation cost modeling.}
Computation models estimate the latency of operators or phases using
analytical formulas, structure-aware simulation, profiles, learned predictors,
or combinations thereof. Roofline-style models estimate latency from FLOPs,
throughput, and efficiency. Structure-aware models additionally represent
hardware hierarchy, tiling, memory movement, and scheduling. Profiling-based
methods reuse measured operator latency. Analytical models can thus provide
broad coverage while profiles calibrate important or hardware-specific
kernels.

\textbf{Communication cost modeling.}
Communication models estimate the latency introduced by distributed
parallelism. Data, tensor, pipeline, context, and expert parallelism induce
collective or point-to-point operations such as All-Reduce, All-Gather,
Reduce-Scatter, Send, and Receive. Their cost depends on message size,
primitive, group size, bandwidth, latency, topology, and overlap with
computation.

Mechanistic models estimate collectives from topology, bandwidth, latency, and
algorithmic decomposition. Trace-driven models replay measured events and
preserve their dependencies; packet-level simulators represent congestion and
routing at higher computational cost; and data-driven predictors learn latency
from message, topology, and configuration features. The appropriate choice
depends on whether the goal is rapid search, workload-specific optimization,
or high-fidelity architectural analysis.

\subsection{Simulation Outputs and Strategy Search}

The predicted iteration time and per-device peak memory provide the
efficiency and feasibility criteria for comparing candidate configurations.
The diagnostic outputs connect these predictions to the operator and
communication measurements used in component-level evaluation.

\textbf{Configuration-selection objective.}
Iteration time and per-device memory can be combined in a task-specific cost
function. Consistent with the mapping
$\mathcal{S}_\theta(M,P,N)\!\to\!(\widehat{T},\widehat{\mathbf{V}},\widehat{C},\widehat{z})$,
we write
\begin{equation}
    \widehat{C}=C(\widehat{T},\widehat{\mathbf{V}},N),
    \label{eq:app-cost-function}
\end{equation}
where $N$ contains the system characteristics. In the benchmark of
Section~\ref{sec:benchmark}, $\widehat{C}=\widehat{T}$: configurations are
ranked by predicted iteration time after the per-device capacity check.
The general formulation also accommodates resource cost or other objectives
when required by an application.

\textbf{Strategy search.}
Configuration search minimizes predicted cost over valid strategies that
pass the predicted feasibility check:
\begin{equation}
    \min_{\substack{P\in\mathcal{K}(M,N)\\
    I_{\mathrm{feas}}(M,P,N)=1}} \widehat{C}.
    \label{eq:app-strategy-search}
\end{equation}
For the DP/TP/PP setting with CP${=}1$ and no expert parallelism,
resource allocation and recomputation satisfy
\begin{equation}
    P_{tp} \cdot P_{pp} \cdot P_{dp} \leq N_{\mathrm{proc}},
    \label{eq:app-parallel-constraint}
\end{equation}
\begin{equation}
    P_{rc} \in \{0,1\},
    \label{eq:app-recompute-constraint}
\end{equation}
Here $P_{tp}$, $P_{pp}$, and $P_{dp}$ are the tensor-, pipeline-, and
data-parallel degrees; $P_{rc}$ is the recomputation switch; and
$N_{\mathrm{proc}}$ is the number of available processors.
With $M$ and $N$ fixed, these choices define the candidate training
configurations. Table~\ref{tab:app-training-strategy} groups the inputs that
specify a simulation instance; the benchmark varies the training dimensions
within each model setting and system.

\begin{table*}[t]
    \centering
    \caption{Configuration inputs and their roles in simulation and search.}
    \label{tab:app-training-strategy}
    \small
    \begin{tabularx}{\linewidth}{@{}>{\RaggedRight\arraybackslash}p{0.23\linewidth}
        >{\RaggedRight\arraybackslash}X>{\RaggedRight\arraybackslash}p{0.27\linewidth}@{}}
        \toprule
        \textbf{Input} & \textbf{Representative parameters} & \textbf{Role} \\
        \midrule
        Model and workload $M$
        & Architecture, layer dimensions, attention heads, sequence length,
          global batch size
        & Defines the training workload. \\
        Training configuration $P$
        & DP/TP/PP/CP/EP, micro-batch size, recomputation, precision,
          optimizer-state partitioning, fusion, and overlap
        & Defines candidate execution strategies for a fixed workload and system. \\
        System configuration $N$
        & Device count, compute capability, memory capacity and bandwidth,
          interconnect, and topology
        & Defines resources and capacity constraints. \\
        \bottomrule
    \end{tabularx}
\end{table*}
\FloatBarrier

\FloatBarrier
\section{LM-SimBench Dataset Details}
\label{app:dataset_details}

This section provides the collection and storage details for the dataset
introduced in Section~\ref{sec:dataset}. Shared model-setting, configuration,
and rank identifiers connect the six measurement views from each training
run. We describe their coverage, collection environment, processing pipeline,
and file schema so that the same records can be used for simulator evaluation
and execution analysis.

\subsection{Workload and Configuration Coverage}
\label{app:configuration_coverage}

Table~\ref{tab:dataset_scale} summarizes the primary NPU collection, and
Table~\ref{tab:appendix_parallel_coverage} gives the per-model coverage. For
each model setting, the latter lists the number of feasible configurations and
the observed degrees of each parallelism dimension. A model setting specifies
a model and its training sequence length; a parallel configuration combines
parallelism degrees and a recomputation setting. Dataset size excludes raw
profiler dumps.

\begin{table*}[t]
\centering
\caption{Scale and coverage of the processed LM-SimBench dataset.}
\label{tab:dataset_scale}
\small
\setlength{\tabcolsep}{5pt}
\renewcommand{\arraystretch}{1.10}
\begin{tabularx}{\linewidth}{p{0.32\linewidth}X}
\toprule
\textbf{Item} & \textbf{Value} \\
\midrule
Model settings & 23 \\
Parallel configurations & 348 \\
Data format & CSV \\
Data modality & Tabular profiling records \\
Model coverage & Dense and MoE models from 10 LLM families \\
Configuration dimensions & DP, TP, PP, CP, EP, recomputation \\
Profiling views & Iteration time, peak memory, time breakdown, operator latency, communication, matrix kernels \\
Processed dataset size & 26.2 GB \\
Raw profiles & 550 GB \\
\bottomrule
\end{tabularx}
\end{table*}

Feasible coverage varies with architecture, parameter count, context length,
memory capacity, and cluster size. Smaller settings therefore admit broader
DP/TP/PP/CP combinations than some larger or more constrained settings. For
MoE workloads, we additionally report expert parallelism because it changes
both computation placement and communication.

\begin{table*}[!htbp]
\centering
\caption{Observed parallelism degrees and configuration counts for each model setting in LM-SimBench.}
\label{tab:appendix_parallel_coverage}
\footnotesize
\setlength{\tabcolsep}{0pt}
\renewcommand{\arraystretch}{1.10}
\begin{tabular*}{\textwidth}{@{\extracolsep{\fill}}lccccccc@{}}
\toprule
\textbf{Model-setting identifier} & \textbf{Configs} & \textbf{DP} & \textbf{TP} & \textbf{PP} & \textbf{CP} & \textbf{EP} & \textbf{Recompute} \\
\midrule
\texttt{gemma2\_27b\_8k} & 3 & 1/2 & 4/8 & 2 & 1/2 & -- & RC \\
\texttt{gemma2\_9b\_8k} & 5 & 1/2 & 2/4 & 2 & 1/2 & -- & RC/noRC \\
\texttt{glm4\_9b\_8k} & 6 & 1/2 & 2 & 2/4 & 1/2 & -- & RC/noRC \\
\texttt{internlm25\_1point8b\_32k} & 16 & 1/2/4 & 1/2/4 & 1/2/4 & 2/4/8 & -- & RC/noRC \\
\texttt{internlm25\_20b\_32k\_16p} & 9 & 1/2 & 2/4/8 & 2/4/8 & 1/2/4 & -- & RC \\
\texttt{internlm25\_7b\_32k} & 5 & 1/2 & 2/4 & 2/4 & 1/2 & -- & RC/noRC \\
\texttt{internlm3\_8b\_8k} & 7 & 1/2 & 1/2 & 2/4 & 2/4 & -- & RC/noRC \\
\texttt{mistral\_7b\_32k} & 11 & 1/2/4 & 1/2/4/8 & 1/2/4/8 & 1 & -- & RC/noRC \\
\texttt{phi35\_mini\_4k} & 38 & 1/2/4/8 & 1/2/4/8 & 1/2/4/8 & 1/2/4 & -- & RC/noRC \\
\texttt{phi35\_moe\_4k} & 10 & 1 & 1/2/4/8 & 1/2/4/8 & 1 & 1/2/4/8 & RC/noRC \\
\texttt{qwen15\_1.8b\_8k} & 40 & 1/2/4/8 & 1/2/4/8 & 1/2/4/8 & 1/2/4/8 & -- & RC/noRC \\
\texttt{qwen15\_4b\_8k} & 40 & 1/2/4/8 & 1/2/4/8 & 1/2/4/8 & 1/2/4/8 & -- & RC/noRC \\
\texttt{qwen25\_0point5\_32k} & 6 & 1/2 & 2 & 2/4 & 1/2 & -- & RC/noRC \\
\texttt{qwen25\_14b\_8k} & 1 & 1 & 4 & 8 & 1 & -- & RC \\
\texttt{qwen25\_32b\_32k} & 4 & 1/2 & 4/8 & 4 & 1/2 & -- & RC/noRC \\
\texttt{qwen25\_3b\_32k} & 6 & 1 & 1/2 & 1/2 & 2/4 & -- & RC/noRC \\
\texttt{qwen25\_72b\_8k} & 3 & 1 & 4/8 & 4/8 & 1 & -- & RC/noRC \\
\texttt{qwen25\_7b\_8k} & 6 & 1/2 & 1/2 & 4 & 1/2 & -- & RC/noRC \\
\texttt{qwen2\_1.5b\_4k} & 26 & 1/2/4/8 & 1/2 & 1/2/4 & 1/2/4 & -- & RC/noRC \\
\texttt{qwen3\_0.6b\_4k} & 38 & 1/2/4/8 & 1/2/4/8 & 1/2/4 & 1/2/4/8 & -- & RC/noRC \\
\texttt{qwen3\_1.7b\_4k} & 38 & 1/2/4/8 & 1/2/4/8 & 1/2/4 & 1/2/4/8 & -- & RC/noRC \\
\texttt{qwen3\_30b\_a3b\_4k} & 2 & 1 & 4 & 2 & 2 & 1 & RC/noRC \\
\texttt{qwen3\_8b\_4k} & 28 & 1/2/4 & 1/2/4/8 & 1/2/4 & 1/2/4 & -- & RC/noRC \\
\bottomrule
\end{tabular*}
\end{table*}

The release identifiers in Table~\ref{tab:appendix_parallel_coverage}
encode the model and training sequence length. In the case studies,
\texttt{qwen15\_1.8b\_8k} denotes Qwen1.5-1.8B at 8K tokens, and
\texttt{qwen25\_32b\_32k} denotes Qwen2.5-32B at 32K tokens.
The cross-platform identifiers \texttt{qwen3\_0.6b\_4k} and
\texttt{qwen3\_1.7b\_4k} denote Qwen3-0.6B and Qwen3-1.7B,
respectively, both at 4K tokens.

The cross-platform supplement pairs V100/Megatron-LM measurements with a
newly collected 910B3/MindSpeed-LLM reference for two Qwen3 workloads. Its
candidate names match a CP${=}1$ subset of the primary collection; its paired
measurements and comparisons are documented in
Appendix~\ref{app:cross_platform}.

\subsection{Profiling Platform and Software Stack}
\label{app:profiling_platform}

The primary collection uses Ascend 910B3 NPUs with MindSpeed-LLM.
Table~\ref{tab:appendix_platform_stack} records the host, accelerator,
interconnect, and software versions used for these runs.

\begin{table*}[t]
\centering
\caption{Hardware and software environment used to collect the primary
LM-SimBench profiles.}
\label{tab:appendix_platform_stack}
\small
\renewcommand{\arraystretch}{1.15}
\setlength{\tabcolsep}{5pt}
\begin{tabularx}{\linewidth}{p{0.24\linewidth}p{0.28\linewidth}X}
\toprule
\textbf{Category} & \textbf{Item} & \textbf{Value} \\
\midrule
\multirow{3}{*}{Host processor}
& CPU architecture & aarch64 \\
& CPU model & Kunpeng-920 \\
& CPU cores & 192 \\
\midrule
\multirow{3}{*}{Accelerator}
& NPU model & Ascend 910B3 \\
& HBM per NPU & 65536 MiB \\
& NPU interconnect & HCCS \\
\midrule
\multirow{3}{*}{Ascend software stack}
& CANN toolkit & 8.0.0 \\
& CANN kernels & 910B, 8.0.0 \\
& CANN Neural Network Acceleration Library (NNAL) & 8.0.0 \\
\midrule
\multirow{4}{*}{Deep learning runtime}
& Python & 3.9.11 \\
& PyTorch & 2.1.0 \\
& torch-npu & 2.1.0.post3 \\
& Transformers & 4.43.2 \\
\midrule
\multirow{2}{*}{Training framework}
& MindSpeed & 0.8.0 \\
& MindSpeed-LLM & branch 2.1.0, commit \texttt{bf1e61f4} \\
\bottomrule
\end{tabularx}
\end{table*}

\subsection{Processing Pipeline and Quality Control}
\label{app:data_processing_pipeline}

Algorithm~\ref{alg:data_processing} converts per-rank Ascend Profiler
outputs into the six views in Table~\ref{tab:data_views_tasks}. Peak memory,
time breakdowns, operator latencies, communication events, and matrix-kernel
records are extracted from the corresponding profiler files.
Configuration-level iteration time is derived by summing the profiler-reported
\texttt{Stage} and \texttt{Bubble} durations for each rank and taking the
maximum across ranks, with microseconds converted to milliseconds.
Shared model-setting, configuration, and rank identifiers link the
configuration-level labels to their underlying execution records.

\begin{algorithm}[H]
\small
\caption{Data processing pipeline for LM-SimBench}
\label{alg:data_processing}
\begin{algorithmic}[1]
\REQUIRE Raw profiler directory $\mathcal{P}_{\mathrm{raw}}$
\ENSURE Structured dataset directory $D$
\STATE Initialize the structured output directory $D$
\FOR{each model $m$ in $\mathcal{P}_{\mathrm{raw}}$}
    \STATE Identify feasible parallel configurations of $m$
    \FOR{each configuration $c$ of $m$}
        \STATE Parse DP, TP, PP, CP, EP, and recomputation settings from $c$
        \FOR{each rank directory $r$ under $c$}
            \STATE Read profiler metadata and obtain the rank identifier
            \STATE Extract memory, operator, time-breakdown, and communication records
            \STATE Extract matrix-kernel features from kernel-level profiler files
        \ENDFOR
        \STATE Sum \texttt{Stage} and \texttt{Bubble} for each rank; take the maximum across ranks and convert to milliseconds
        \STATE Aggregate rank-level records into structured CSV files
    \ENDFOR
\ENDFOR
\STATE Normalize field names, rank identifiers, units, and directory layout
\STATE Generate manifest records for coverage and size checks
\RETURN Structured dataset directory $D$
\end{algorithmic}
\end{algorithm}

Quality checks cover configuration parsing, rank identifiers, field names,
units, and file coverage. The manifest records configuration and file counts
and storage sizes for consistency checks.

\subsection{File Schema and Field Definitions}
\label{app:file_schema}

The directory hierarchy preserves the associations established during
processing. The root manifest indexes model settings; summary files within
each model-setting directory contain configuration- and rank-level
measurements; and configuration/rank directories retain the detailed
communication and matrix-kernel records:

\noindent\begin{minipage}{\columnwidth}
\begin{verbatim}
LM-SimBench/
  data_manifest.csv
  <model_name>/
    step_time_summary.csv
    peak_memory.csv
    step_trace_time_summary.csv
    operator_latency_by_category.csv
    comm_summary.csv
    <parallel_config>/
      rank_<id>/
        comm_time.csv
        matrix_operator_latency.csv
\end{verbatim}
\end{minipage}
\par\smallskip

Table~\ref{tab:appendix_file_schema_overview} describes each file's location,
granularity, and benchmark role. Table~\ref{tab:appendix_file_fields} lists
representative join keys, targets, and features. The dataset repository
provides complete column definitions.

\begin{table*}[!htbp]
\centering
\caption{
Storage location, profiling granularity, and benchmark role of the main CSV files.
}
\label{tab:appendix_file_schema_overview}
\small
\setlength{\tabcolsep}{5pt}
\renewcommand{\arraystretch}{1.15}
\begin{tabularx}{\textwidth}{>{\RaggedRight\arraybackslash}p{0.24\textwidth}
                                >{\RaggedRight\arraybackslash}p{0.18\textwidth}
                                >{\RaggedRight\arraybackslash}p{0.20\textwidth}
                                >{\RaggedRight\arraybackslash}X}
\toprule
\textbf{File} & \textbf{Location} & \textbf{Granularity} & \textbf{Benchmark role} \\
\midrule
\path{data_manifest.csv}
& Dataset root
& Dataset / model level
& Provides the global index of model directories, configuration counts, rank-directory counts, file coverage, and dataset-scale statistics. \\

\midrule
\path{step_time_summary.csv}
& Model directory
& Configuration level
& Provides configuration-level iteration-time labels derived from profiler time breakdowns for runtime prediction and configuration ranking. \\

\midrule
\path{peak_memory.csv}
& Model directory
& Rank level
& Provides rank-level peak memory labels for memory modeling, feasibility analysis, and imbalance analysis across ranks. \\

\midrule
\path{step_trace_time_summary.csv}
& Model directory
& Rank level
& Decomposes iteration time into computation, communication, idle time, and pipeline-related components for bottleneck diagnosis. \\

\midrule
\path{operator_latency_by_category.csv}
& Model directory
& Operator-category / rank level
& Aggregates profiler operators into semantic categories for operator-latency modeling and module-level performance analysis. \\

\midrule
\path{comm_summary.csv}
& Model directory
& Communication-group / rank level
& Provides aggregated communication statistics for comparing collective and point-to-point communication costs across configurations. \\

\midrule
\path{comm_time.csv}
& Configuration / rank directory
& Communication-event / rank level
& Preserves event-level communication behavior for analyzing waiting time, transfer size, link bandwidth, and communication bottlenecks. \\

\midrule
\path{matrix_operator_latency.csv}
& Configuration / rank directory
& Matrix-kernel / rank level
& Provides matrix-operator shapes, latency, data types, and hardware-counter features for GEMM modeling and hardware-aware analysis. \\
\bottomrule
\end{tabularx}
\end{table*}

\begin{table*}[!htbp]
\centering
\caption{
Representative join keys, targets, and features in the main CSV files.
}
\label{tab:appendix_file_fields}
\small
\setlength{\tabcolsep}{5pt}
\renewcommand{\arraystretch}{1.15}
\begin{tabularx}{\textwidth}{>{\RaggedRight\arraybackslash}p{0.26\textwidth}
                                >{\RaggedRight\arraybackslash}X}
\toprule
\textbf{File} & \textbf{Representative fields} \\
\midrule
\path{data_manifest.csv}
& \path{model_name}, \path{parallel_config_count}, \path{rank_directory_count}, \path{size_bytes}, \path{size_human}. \\

\midrule
\path{step_time_summary.csv}
& \path{optimization_config}, \path{iteration_time_ms}. \\

\midrule
\path{peak_memory.csv}
& \path{optimization_config}, \path{rank_id}, \path{peak_memory_mb}. \\

\midrule
\path{step_trace_time_summary.csv}
& \path{optimization_config}, \path{rank_id}, \path{computing}, \path{communication}, \path{communication_not_overlapped}, \path{overlapped}, \path{free}, \path{stage}, \path{bubble}. \\

\midrule
\path{operator_latency_by_category.csv}
& \path{optimization_config}, \path{rank_id}, \path{op_category}, \path{op_type}, \path{core_type}, \path{count}, \path{total_time_us}, \path{avg_time_us}, \path{ratio_percent}. \\

\midrule
\path{comm_summary.csv}
& \path{comm_scope}, \path{comm_type}, \path{comm_group}, \path{op_count}, \path{elapsed_time_sum_ms}, \path{wait_time_sum_ms}, \path{link_bandwidth}. \\

\midrule
\path{comm_time.csv}
& \path{comm_scope}, \path{comm_type}, \path{comm_group}, \path{op_name}, \path{elapsed_time_ms}, \path{wait_time_ms}, \path{link_type}, \path{link_transit_size_mb}, \path{link_bandwidth_gbps}. \\

\midrule
\path{matrix_operator_latency.csv}
& \path{type}, \path{duration_us}, \path{input_shapes}, \path{output_shapes}, \path{a_shape}, \path{b_shape}, \path{c_shape}, \path{batch}, \path{m}, \path{n}, \path{k}, \path{aic_mac_ratio}, \path{cube_utilization_percent}. \\
\bottomrule
\end{tabularx}
\end{table*}

Within a model setting, \path{optimization_config} joins configuration-level
records and \path{rank_id} identifies the corresponding rank-level views.
Dense-model configuration names follow
\path{dp<dp>-tp<tp>-pp<pp>-cp<cp>-<RC|noRC>}; MoE names may additionally
contain \path{ep<ep>}. The suffixes \path{RC} and \path{noRC} specify the
recomputation setting. These keys connect the iteration-time label to the
rank-level summaries and the event- and kernel-level files for the same run.

\subsection{Documentation, Licensing, and Maintenance}
\label{app:dataset_documentation}
\label{app:dataset_availability}

\paragraph{Documentation.}
The resources listed at the beginning of this supplement include dataset
cards, field descriptions, usage examples, and Croissant metadata.
Table~\ref{tab:appendix_dataset_documentation} summarizes the data's use and
provenance.

\begin{table*}[!htbp]
\centering
\caption{Use, coverage, and provenance of LM-SimBench.}
\label{tab:appendix_dataset_documentation}
\small
\setlength{\tabcolsep}{5pt}
\renewcommand{\arraystretch}{1.15}
\begin{tabularx}{\textwidth}{>{\RaggedRight\arraybackslash}p{0.24\textwidth}
                                >{\RaggedRight\arraybackslash}X}
\toprule
\textbf{Topic} & \textbf{Summary} \\
\midrule
Intended use
& Training-simulator evaluation and development: time and memory prediction,
configuration ranking, operator and communication analysis, and calibration. \\
\midrule
Coverage and evaluation scope
& Measurements describe the sampled models, feasible configurations, and
software/hardware stacks in Section~\ref{app:configuration_coverage} and
Appendix~\ref{app:cross_platform}. Coverage depends on model size, memory
capacity, and platform implementations. Serving behavior and model-quality,
safety, or fairness assessment require separate data. \\
\midrule
Provenance and record type
& Controlled distributed training runs produce the logs and profiler outputs;
the processed tables are normalized derivatives of these measured records,
without synthetic profiling samples or human annotations. \\
\midrule
Personal or sensitive information
& The release contains system telemetry, with no intentional personal records,
user-generated text, or private user data. \\
\bottomrule
\end{tabularx}
\end{table*}

\paragraph{License.}
The datasets, metadata, and documentation are released under CC BY 4.0;
the preprocessing and benchmark code is released under Apache-2.0.

\paragraph{Maintenance.}
The release will be versioned, with previous versions retained for
reproducibility and major updates recorded in a changelog. File, schema,
preprocessing, and data-loading issues will be tracked in the code repository.

\FloatBarrier

\clearpage
\section{Simulator Evaluation and Comparison}
\label{app:cross_platform_vtrain}

This section provides the full results supporting
Sections~\ref{subsec:simulator-comparison} and~\ref{subsec:diagnostic-results}.
Calculon and vTrain~\cite{bang2024vtrainsimulationframeworkevaluating}
illustrate how LM-SimBench evaluates analytical and profiling-based
simulators through common time and ranking targets. Intermediate profiles
then connect their overall predictions to computation, memory, and
communication behavior.

\subsection{Evaluation Setup}

We use Qwen1.5-1.8B at an 8K-token sequence length with CP${=}1$ on the
primary NPU stack. Calculon uses analytical cost models; vTrain combines
an execution graph with LM-SimBench profiles supplied through an adapter.
Ten configurations calibrate vTrain, and ten disjoint configurations form
the test set. The shared-task comparison uses the six test configurations
for which both simulators produce valid predictions.

Predicted and measured iteration times are matched by configuration.
Accuracy, ranking, and shortlist metrics follow
Section~\ref{sec:benchmark}; simulation cost is the median software
wall-clock time per configuration in the same CPU and software environment.
The component and memory analyses below use their corresponding profiling
targets, with coverage specified alongside the results.

\subsection{Prediction, Selection, and Simulation Cost}

Table~\ref{tab:vtrain-comparison} expands the comparison in
Table~\ref{tab:main-simulators}. vTrain achieves 1.86\% iteration-time sMAPE
and Spearman $\rho=1.000$, compared with 8.88\% and 0.943 for Calculon.
Its absolute time error is lower on five of the six configurations.
Both simulators identify the measured fastest configuration and recover
the measured top-three set, giving zero Top-1 selection regret.

Calculon's median simulation time is 1.78\,ms per configuration, compared
with 25.46\,ms for vTrain, a $14.3\times$ speedup. Together, the prediction,
selection, and timing metrics characterize how the two modeling approaches
support configuration search: prediction accuracy describes the estimated
runtime, while ranking and regret describe the resulting choice.

\begin{table*}[t]
\centering
\small
\caption{Shared-task evaluation of Calculon and vTrain on six common test
configurations of Qwen1.5-1.8B (8K tokens, CP${=}1$).}
\label{tab:vtrain-comparison}
\setlength{\tabcolsep}{4pt}
\begin{tabularx}{\linewidth}{@{}l>{\RaggedRight\arraybackslash}X>{\RaggedRight\arraybackslash}p{0.24\linewidth}>{\RaggedRight\arraybackslash}p{0.24\linewidth}@{}}
\toprule
Level & Task and metric & Calculon & vTrain \\
\midrule
System & Iteration time, MAE / sMAPE / $R^2$ & 1.059\,s / 8.88\% / 0.630 & \textbf{0.207\,s / 1.86\% / 0.990} \\
Strategy & Iteration-time ranking, Spearman $\rho$ & 0.943 & \textbf{1.000} \\
         & Iteration-time ranking, Kendall $\tau$ & 0.867 & \textbf{1.000} \\
         & Top-1 hit / selection regret & 1 / 0\% & 1 / 0\% \\
         & Top-3 set overlap & 3/3 & 3/3 \\
Efficiency & Median wall-clock per configuration & \textbf{1.78\,ms} & 25.46\,ms \\
\bottomrule
\end{tabularx}
\end{table*}

\subsection{Calibration Coverage}

A separate analysis evaluates how the configuration coverage of the
calibration profiles affects vTrain's predictions. The test set is fixed
at eight configurations. Six TP-endpoint calibration profiles yield
11.77\% sMAPE; ten calibration profiles spanning TP/DP/PP reduce it to
2.08\%. The 9.69-percentage-point reduction illustrates the value of
covering multiple parallelism dimensions when supplying measured costs to
a profiling-based simulator.

\subsection{Diagnosis with Intermediate Profiles}

\paragraph{Calculon: memory and matrix operators.}
Table~\ref{tab:calculon-benchmark-summary} summarizes Calculon's diagnostic
results. Its memory model retains 13 of the 20 measured-feasible
configurations. Time and peak-memory errors and iteration-time rankings
are evaluated on those 13 configurations; memory uses the maximum across
ranks. Matrix-operator predictions are evaluated against per-call kernel
latencies at matching matrix dimensions.

The 6.92\,GiB peak-memory MAE and 43.34\% matrix-operator sMAPE connect the
end-to-end assessment to specific modeling components.
Appendix~\ref{sec:case_study_single_node_small_model} shows the corresponding
time and memory predictions, and
Appendix~\ref{app:detailed_matrix_errors} reports matrix errors by
configuration and shape. These linked views allow a reader to follow an
aggregate metric to the measurements that explain it.

\begin{table*}[t]
\centering
\caption{Calculon diagnostics on Qwen1.5-1.8B (8K tokens, CP${=}1$).
Time, memory, and ranking use the 13 configurations retained from 20
measured-feasible candidates.}
\label{tab:calculon-benchmark-summary}
\small
\setlength{\tabcolsep}{5pt}
\renewcommand{\arraystretch}{1.12}
\begin{tabular*}{\textwidth}{@{\extracolsep{\fill}}llll@{}}
\toprule
\textbf{Task} & \textbf{Level} & \textbf{Metrics} & \textbf{Results} \\
\midrule
Iteration time prediction & System & MAE / sMAPE / $R^2$ & 1.97\,s / 15.96\% / 0.358 \\
Peak memory prediction & System & MAE / sMAPE / $R^2$ & 6.92\,GiB / 39.34\% / $-3.087$ \\
Matrix-operator latency & Component & MAE / sMAPE / $R^2$ & 0.109\,ms / 43.34\% / 0.817 \\
Feasible-configuration retention & System & Retention rate & 65.0\% (13/20) \\
Iteration time ranking & Strategy & Spearman $\rho$ / Top-1 hit & 0.753 / 1 \\
\bottomrule
\end{tabular*}
\par\smallskip
\begin{minipage}{\textwidth}
\footnotesize
Iteration-time and peak-memory prediction and ranking use the 13 configurations
retained by Calculon's memory model; the retention rate uses all 20
measured-feasible configurations. Results follow the order of the listed metrics.
\end{minipage}
\end{table*}

\begin{table*}[t]
\centering
\small
\caption{vTrain component diagnostics on Qwen1.5-1.8B (8K tokens).
Predictions are compared with stage-level profiling targets.}
\label{tab:vtrain-stage-diagnostics}
\setlength{\tabcolsep}{6pt}
\begin{tabular*}{\textwidth}{@{\extracolsep{\fill}}lrrr@{}}
\toprule
\textbf{Target} & \textbf{MAE} & \textbf{sMAPE (\%)} & $\boldsymbol{R^2}$ \\
\midrule
Stage computation time & 716.10\,ms & 11.60 & 0.462 \\
Recomputation-stage time & 37.99\,ms & 2.75 & 0.801 \\
Non-overlapped communication time & 8.710\,s & 136.20 & $-0.643$ \\
\bottomrule
\end{tabular*}
\end{table*}

\paragraph{vTrain: computation and communication stages.}
Stage-level profiles provide evaluation targets for the intermediate costs
emitted by vTrain (Table~\ref{tab:vtrain-stage-diagnostics}). Computation and
recomputation achieve 11.60\% and 2.75\% sMAPE, respectively.
Non-overlapped communication has 136.20\% sMAPE: vTrain's serialized
communication estimates differ from the overlap observed in the training
profiles. The stage breakdown thus identifies communication scheduling as
a concrete component to refine, alongside the accurate recomputation
estimates.

\FloatBarrier
\section{Profiling Cases and Simulator Diagnostics}
\label{app:representative_profiling_cases}

The cases in this section expand the intermediate-profile analysis in
Section~\ref{subsec:diagnostic-results}. Qwen1.5-1.8B connects configuration
selection to time and memory predictions, while Qwen2.5-32B shows how
rank-level profiles reveal the computation and communication behind a
multi-node training iteration. In both cases, the linked views connect
overall behavior to specific ranks, communication primitives, and matrix
operators.

\subsection{Single-Node Case: Qwen1.5-1.8B}
\label{sec:case_study_single_node_small_model}

\begin{figure*}[!htbp]
  \centering
  \includegraphics[width=\textwidth]{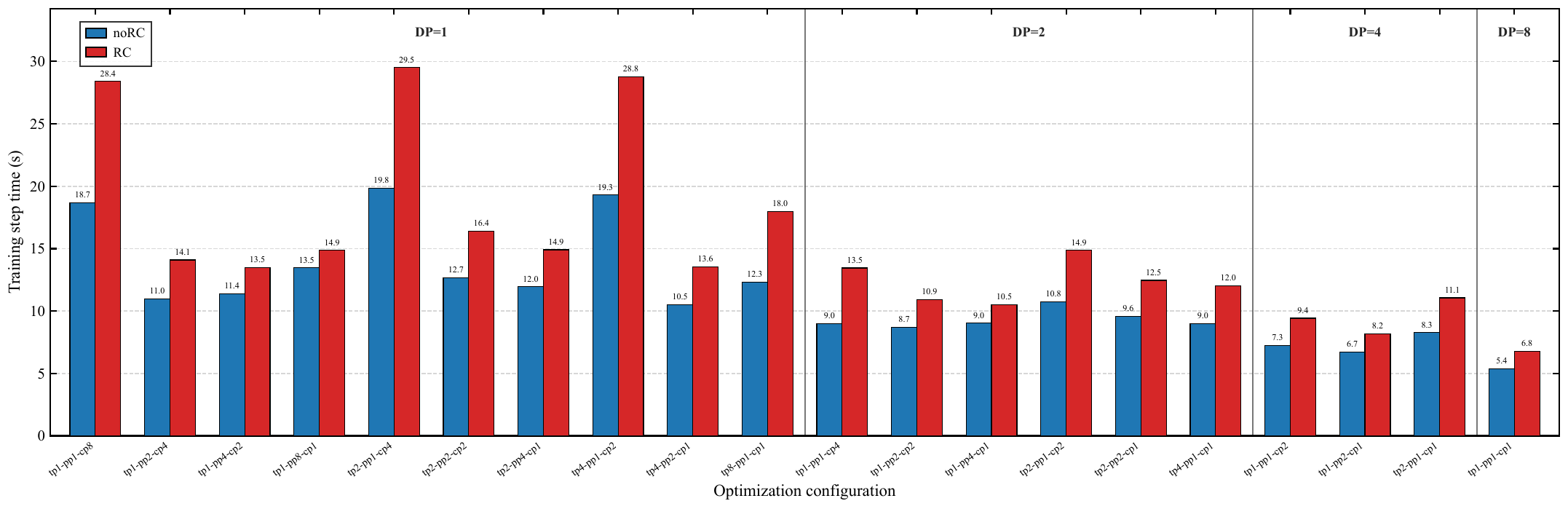}
  \caption{
  \textbf{Iteration time across parallel configurations.}
  Mean iteration time of Qwen1.5-1.8B (8K tokens) under different
  DP/TP/PP/CP and recomputation settings.
  }
  \label{fig:case1_step_time}
\end{figure*}

\begin{figure*}[!htbp]
  \centering
  \includegraphics[width=\textwidth]{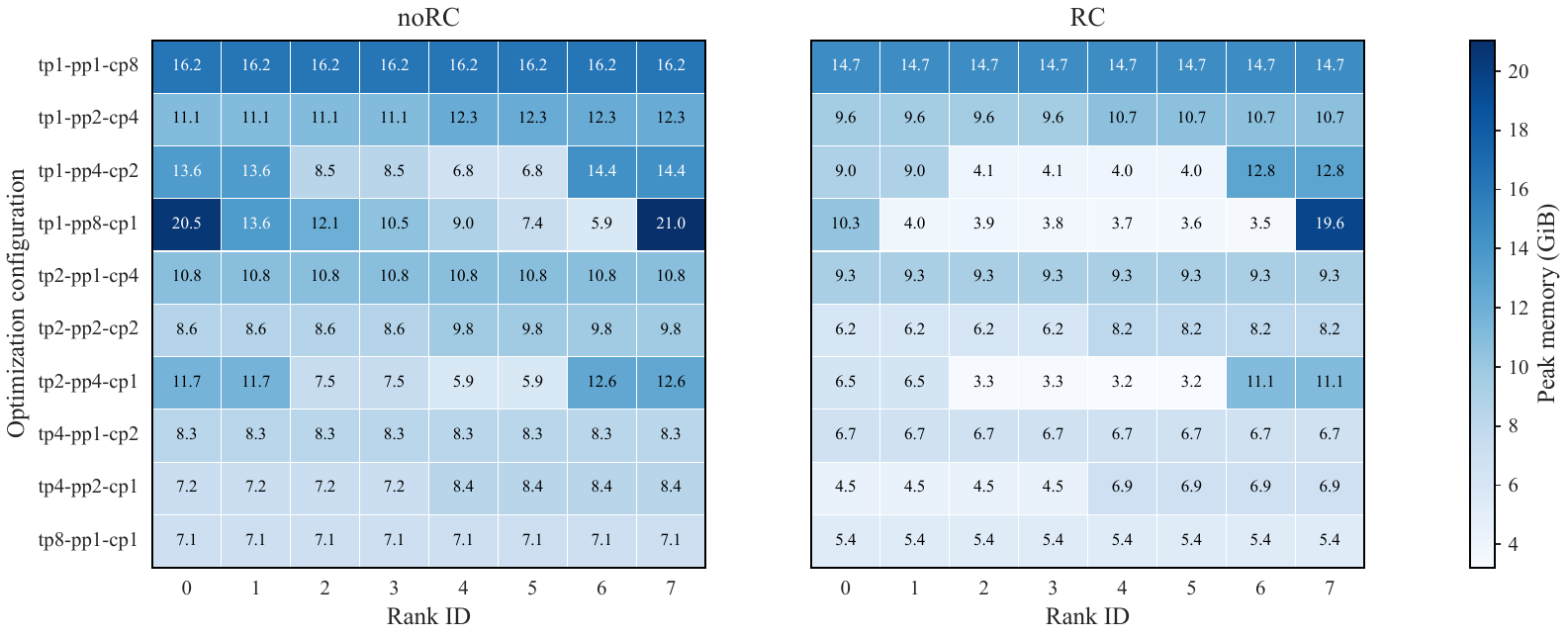}
  \caption{
  \textbf{Rank-level peak memory under DP=1 configurations.}
  Peak memory of Qwen1.5-1.8B (8K tokens) by rank across TP/PP/CP
  configurations.
  }
  \label{fig:case1_peak_memory}
\end{figure*}

\paragraph{Configuration-level time and rank-level memory.}
For Qwen1.5-1.8B at a training sequence length of 8K tokens, varying
DP, TP, PP, CP, and recomputation produces substantial differences in iteration time
(Figure~\ref{fig:case1_step_time}). Rank-level memory measurements expose
how parallel partitioning and recomputation change the memory distribution
within the same workload (Figure~\ref{fig:case1_peak_memory}).

\begin{figure*}[!htbp]
  \centering
  \includegraphics[width=\textwidth]{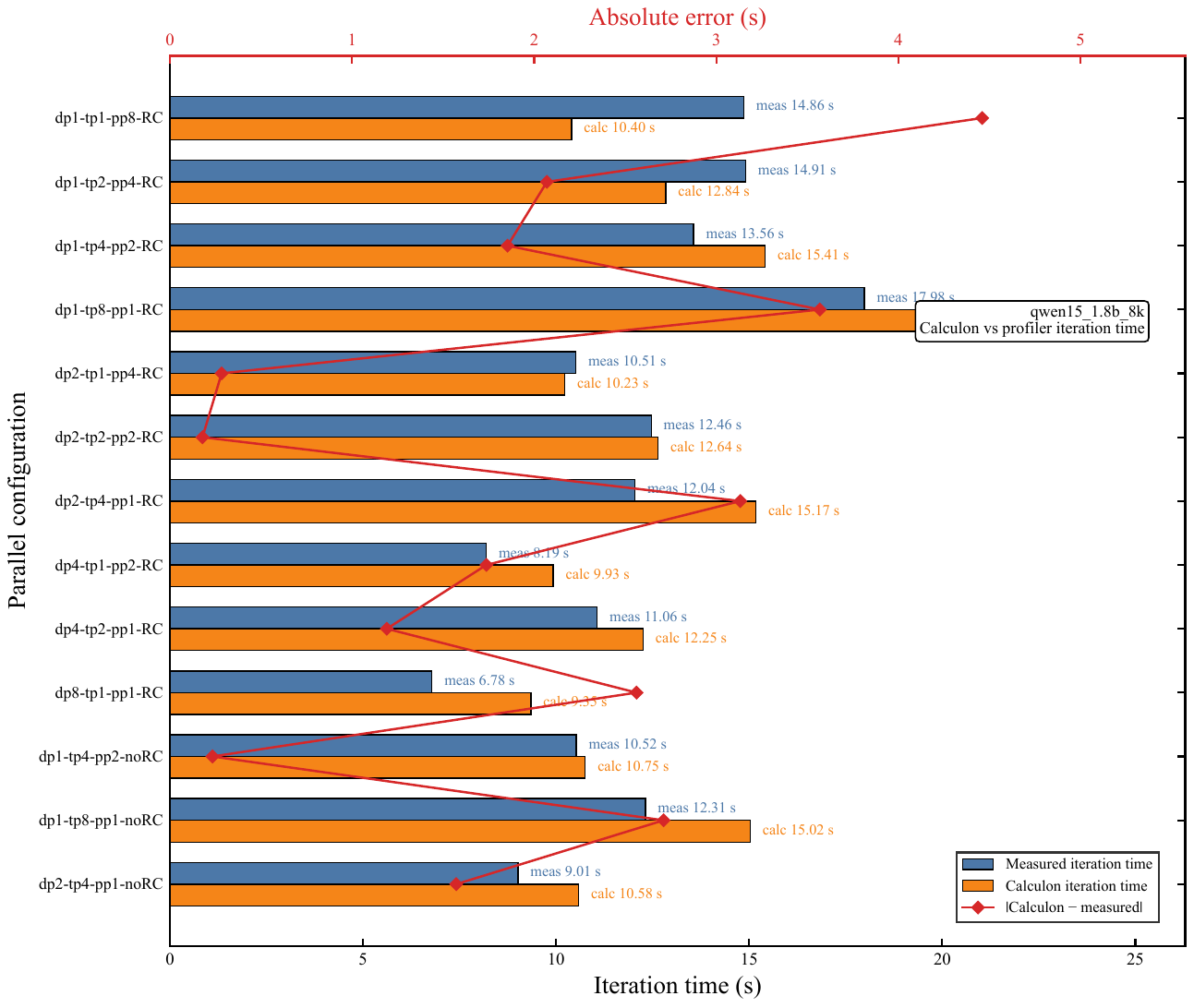}
  \caption{
  \textbf{Calculon iteration-time prediction on Qwen1.5-1.8B (8K tokens).}
  Measured and Calculon-predicted iteration time for the 13 configurations
  accepted by Calculon's memory model.
  }
  \label{fig:calculon-iteration-qwen15}
\end{figure*}

\begin{figure*}[!htbp]
  \centering
  \includegraphics[width=\textwidth]{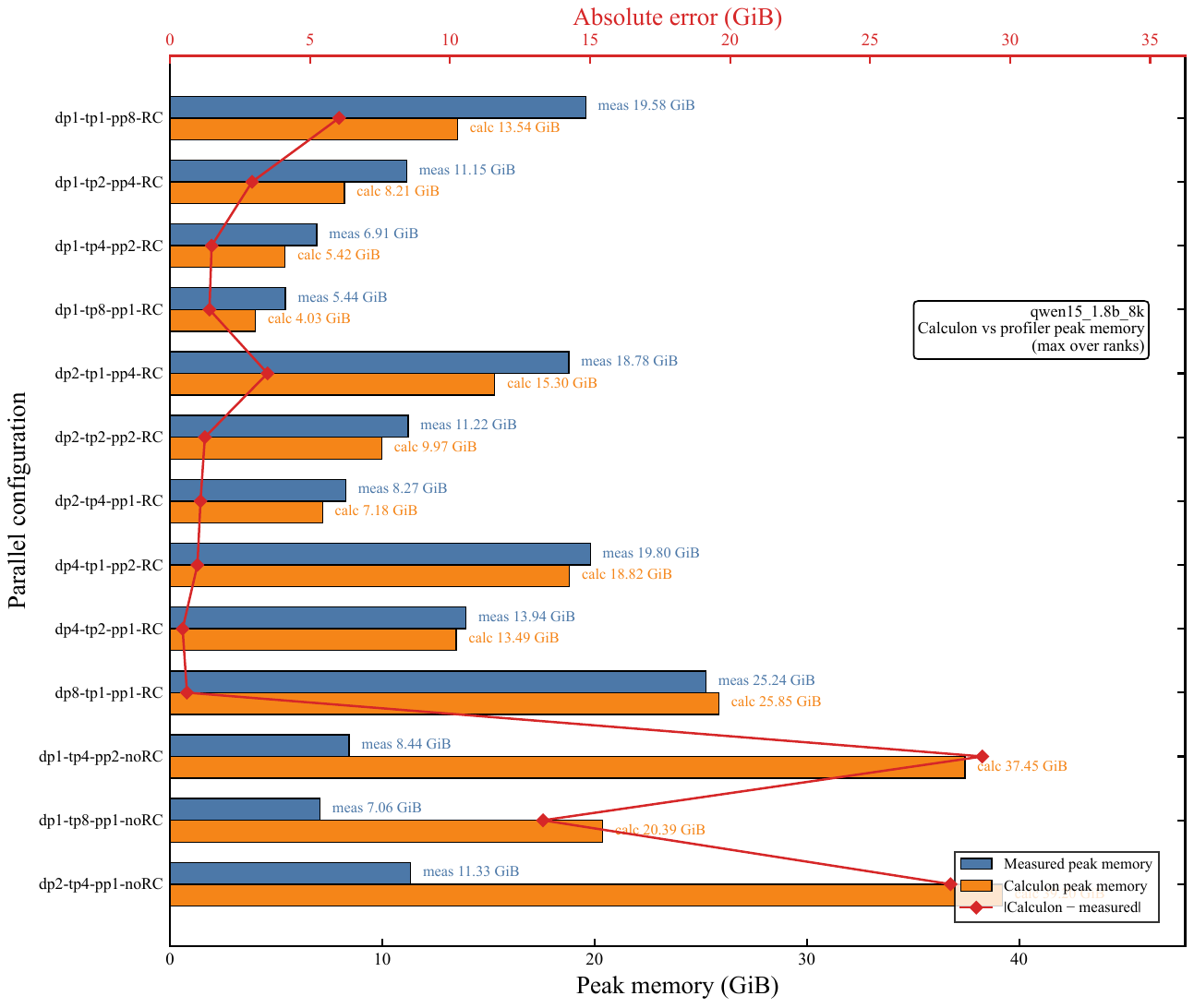}
  \caption{
  \textbf{Calculon peak-memory prediction on Qwen1.5-1.8B (8K tokens).}
  Measured maximum rank-level peak memory and Calculon predictions for 13
  configurations.
  }
  \label{fig:calculon-memory-qwen15}
\end{figure*}

\paragraph{Connecting memory screening to prediction accuracy.}
On the CP${=}1$ subset, Calculon's memory model accepts 13 of the 20
measured-feasible configurations and rejects seven. Its largest memory
overestimates occur without recomputation. Figures~\ref{fig:calculon-iteration-qwen15}
and~\ref{fig:calculon-memory-qwen15} compare time and memory predictions for
the accepted configurations. Together, these measurements distinguish
runtime prediction errors from the loss of feasible candidates during memory
screening. The aggregate metrics are reported in
Table~\ref{tab:calculon-benchmark-summary}; detailed matrix-operator errors
for this workload are collected in
Appendix~\ref{app:detailed_matrix_errors}.

\FloatBarrier

\subsection{Multi-Node Case: Qwen2.5-32B}
\label{sec:case_study_multi_node_dense_model}

\begin{figure*}[!htbp]
  \centering
  \includegraphics[width=\textwidth]{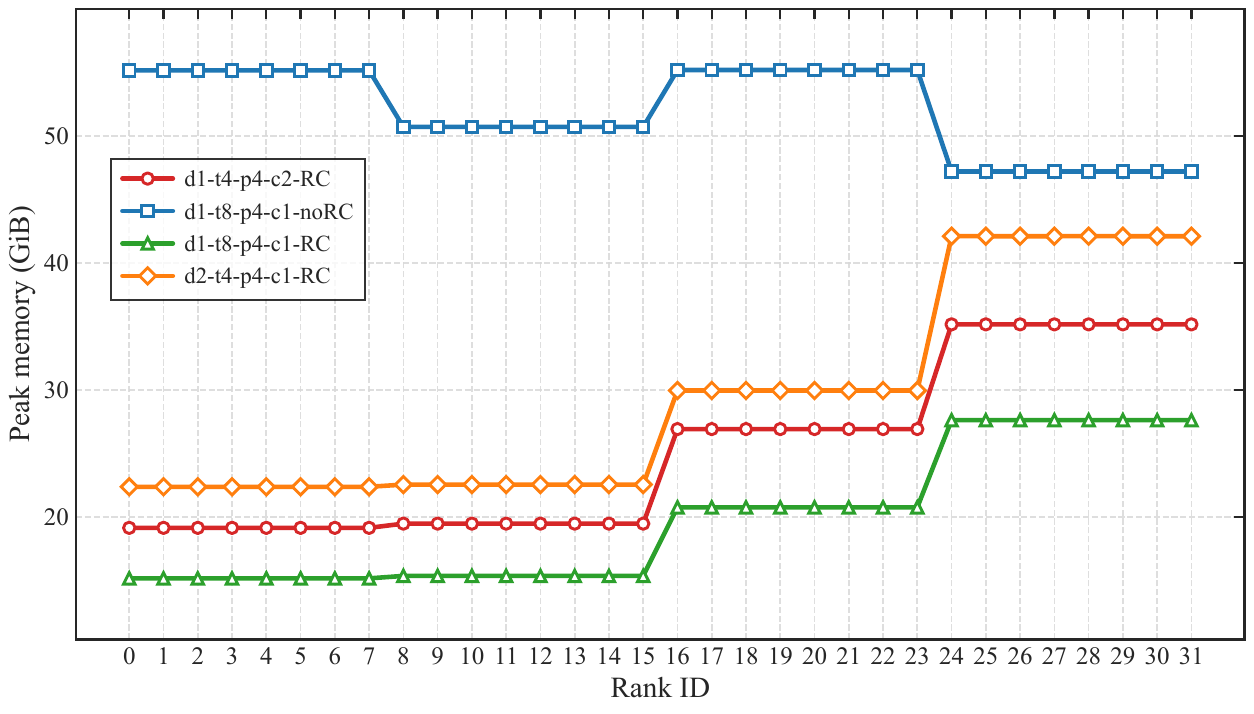}
  \caption{
  \textbf{Rank-level peak memory across configurations.}
  Peak memory by rank for Qwen2.5-32B (32K tokens) under four parallel
  configurations.
  }
  \label{fig:qwen25_32b_peak_memory}
\end{figure*}

\paragraph{Locating costs across ranks.}
For Qwen2.5-32B at a training sequence length of 32K tokens, rank-level
profiles reveal nonuniform memory,
computation, and communication across the distributed execution. Memory
curves vary with parallel partitioning and pipeline-stage assignment
(Figure~\ref{fig:qwen25_32b_peak_memory}); time breakdowns identify the
computation and non-overlapped communication associated with each rank
(Figure~\ref{fig:qwen25_32b_step_breakdown}).

\begin{figure*}[p]
\centering
\begin{subfigure}[t]{0.96\textwidth}
      \centering
      \includegraphics[width=\linewidth]{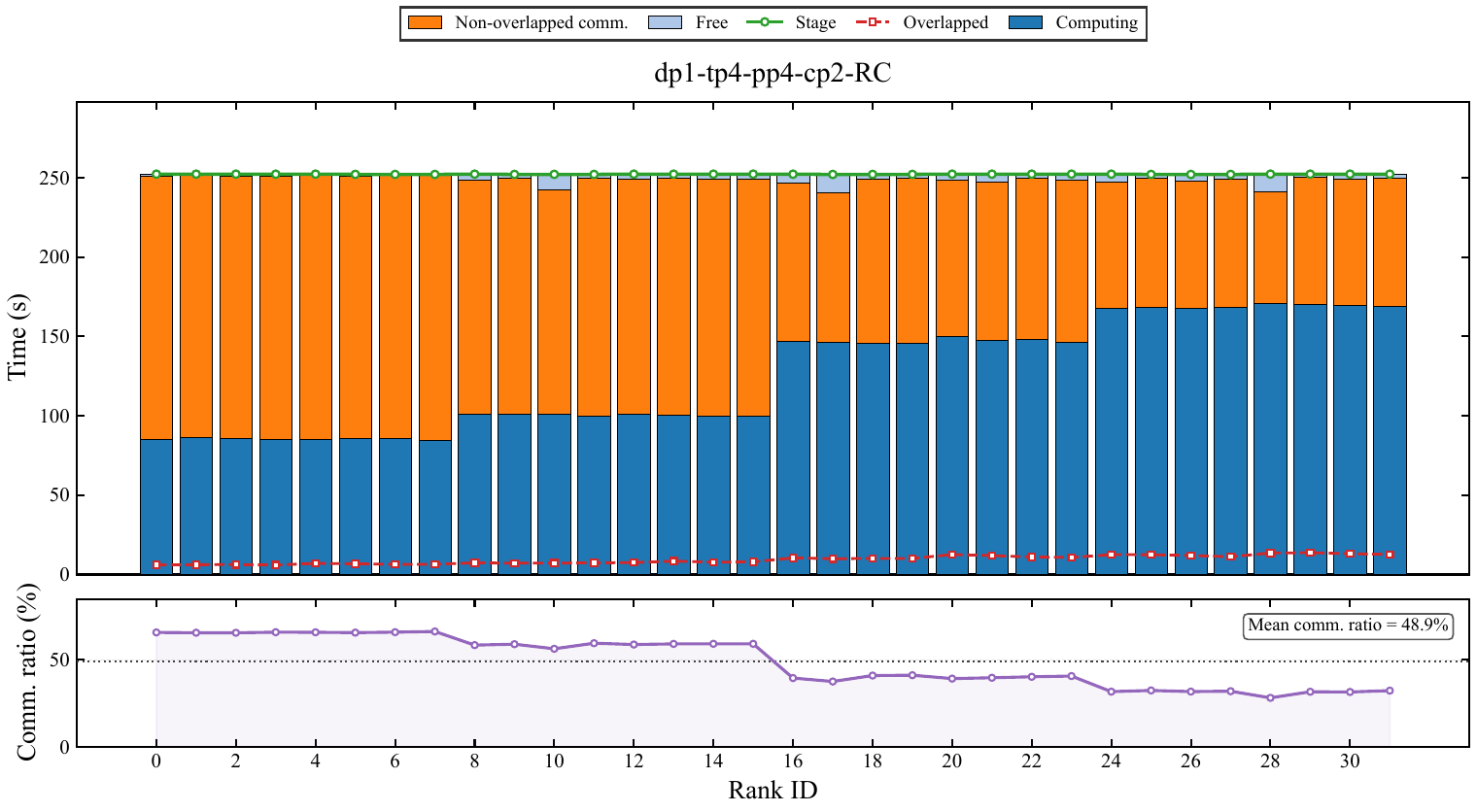}
      \caption{\texttt{dp1-tp4-pp4-cp2-RC}.}
      \label{fig:qwen25_32b_step_breakdown_a}
  \end{subfigure}
\par\medskip
\begin{subfigure}[t]{0.96\textwidth}
      \centering
      \includegraphics[width=\linewidth]{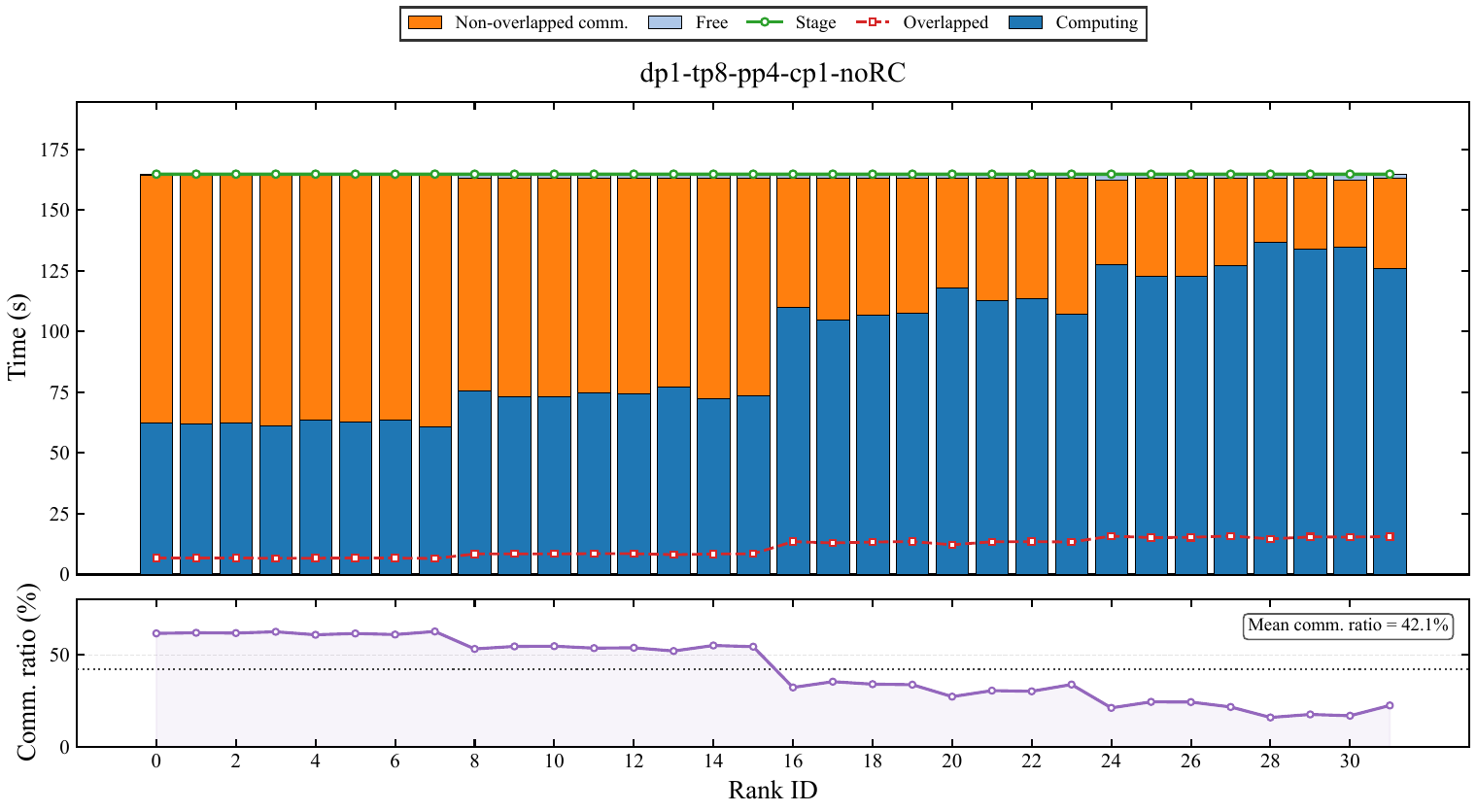}
      \caption{\texttt{dp1-tp8-pp4-cp1-noRC}.}
      \label{fig:qwen25_32b_step_breakdown_b}
  \end{subfigure}
\caption{
  \textbf{Rank-level iteration-time breakdowns.}
  Each panel breaks down iteration time for Qwen2.5-32B (32K tokens) into computation,
  non-overlapped communication, free, stage, and overlapped time and reports
  the communication ratio for one configuration.
  }
  \label{fig:qwen25_32b_step_breakdown}
\end{figure*}

\begin{figure*}[p]
\ContinuedFloat
\centering
\begin{subfigure}[t]{0.96\textwidth}
      \centering
      \includegraphics[width=\linewidth]{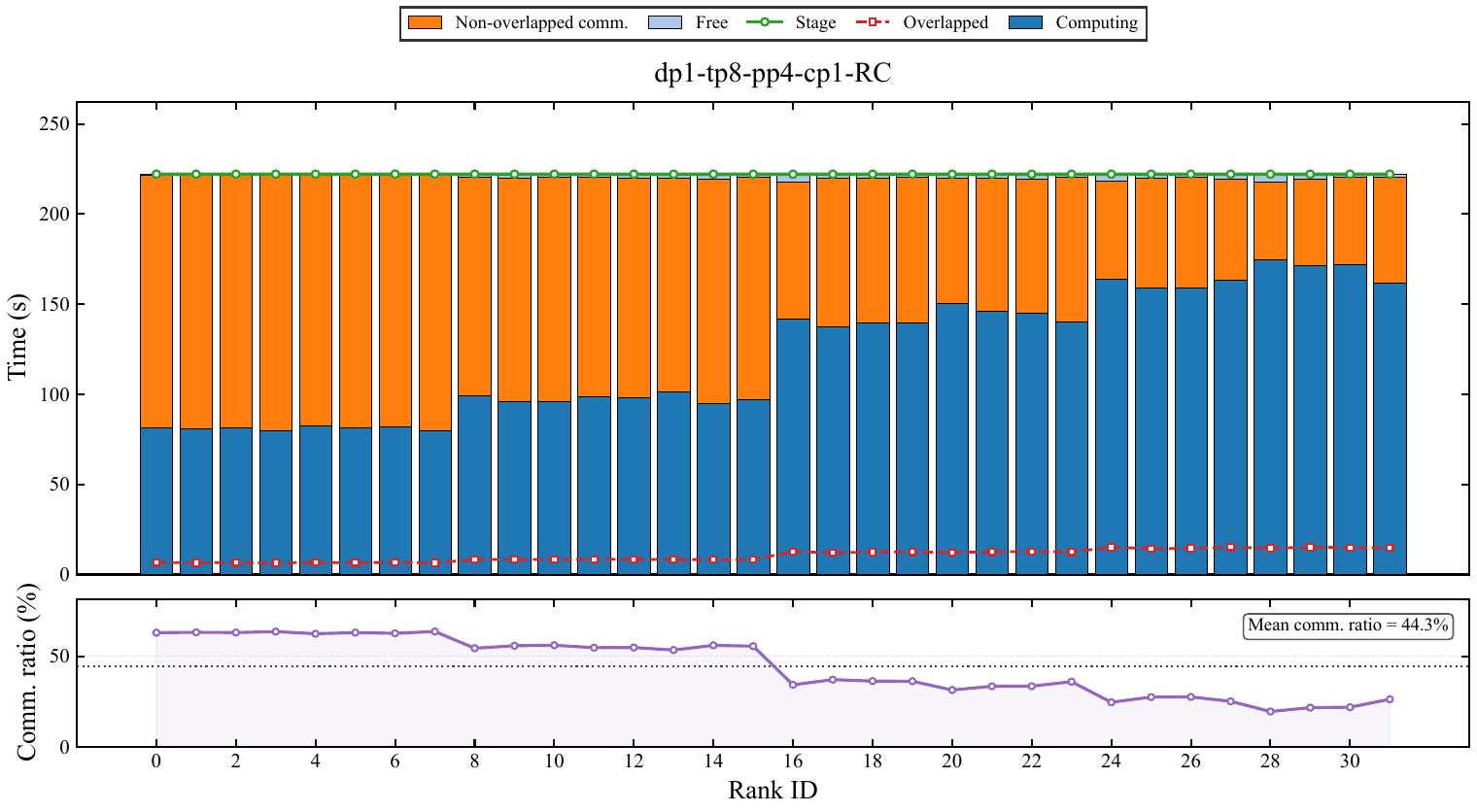}
      \caption{\texttt{dp1-tp8-pp4-cp1-RC}.}
      \label{fig:qwen25_32b_step_breakdown_c}
  \end{subfigure}
\par\medskip
\begin{subfigure}[t]{0.96\textwidth}
      \centering
      \includegraphics[width=\linewidth]{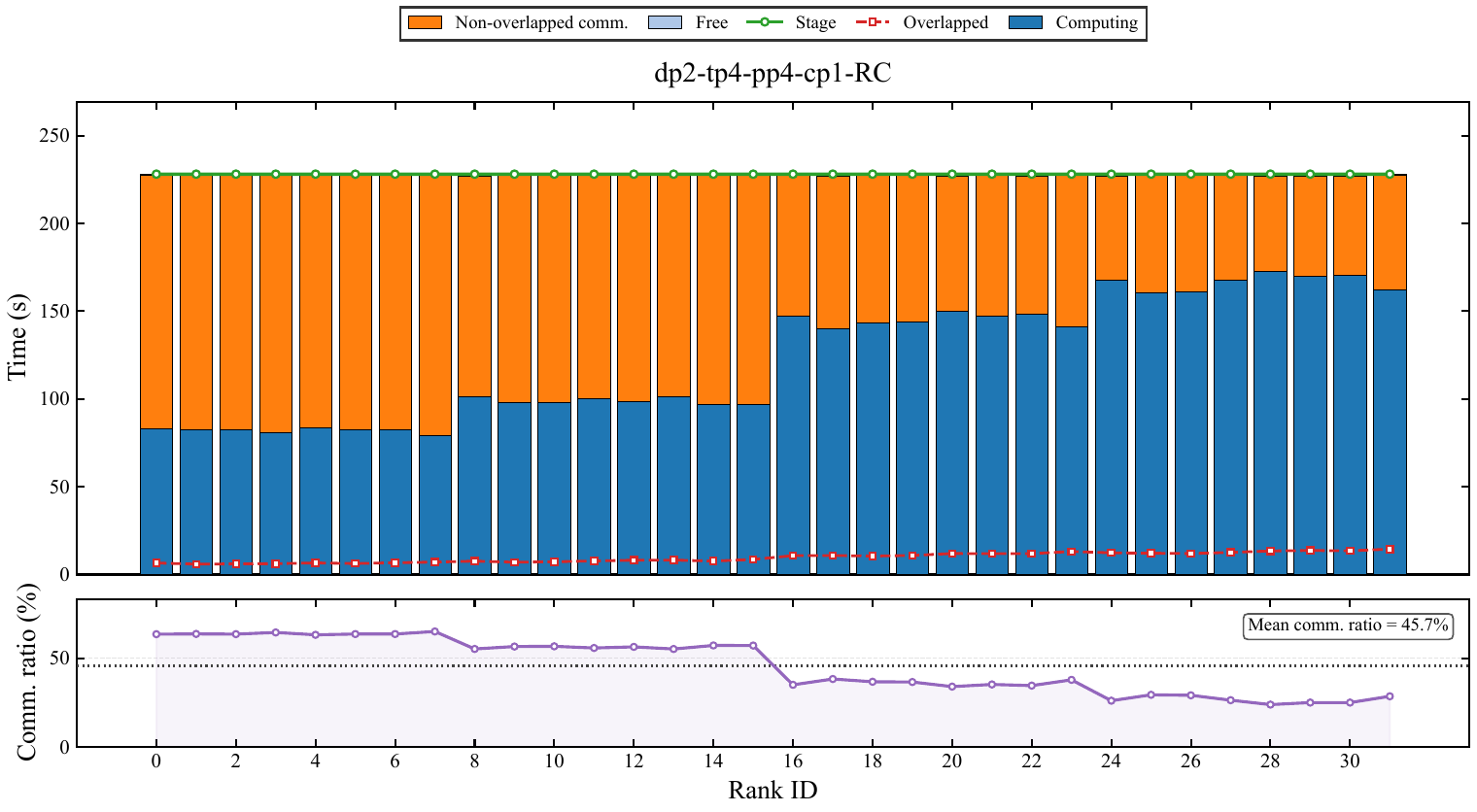}
      \caption{\texttt{dp2-tp4-pp4-cp1-RC}.}
      \label{fig:qwen25_32b_step_breakdown_d}
  \end{subfigure}
\caption{
  \textbf{Rank-level iteration-time breakdowns.}
  Each panel breaks down iteration time for Qwen2.5-32B (32K tokens) into computation,
  non-overlapped communication, free, stage, and overlapped time and reports
  the communication ratio for one configuration.
  }
  
\end{figure*}

\begin{figure*}[!htbp]
  \centering
  \includegraphics[width=0.96\textwidth]{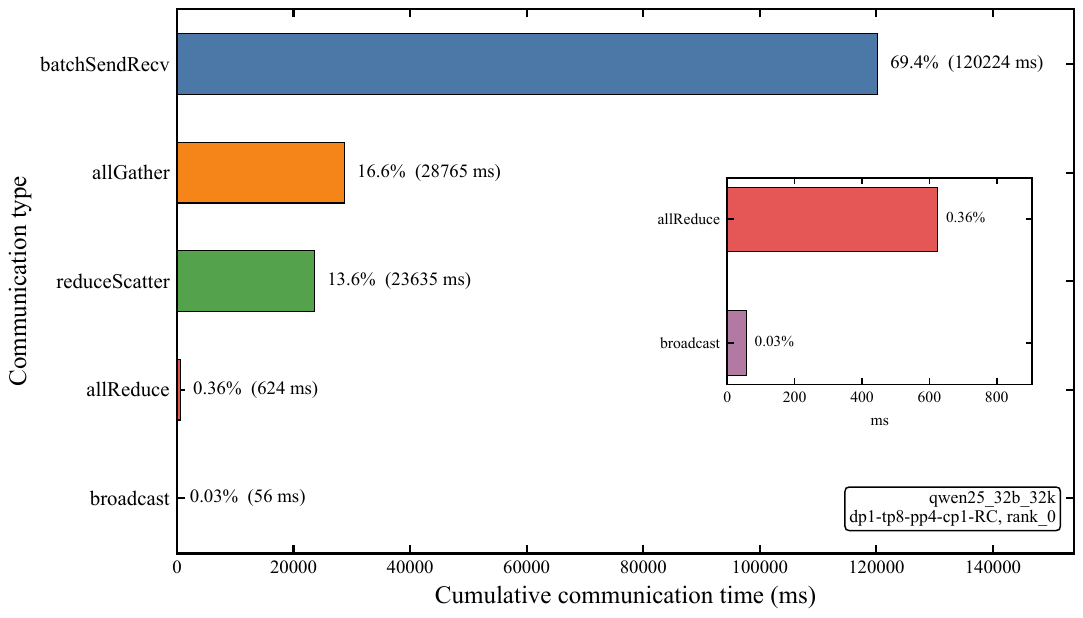}
  \caption{
  \textbf{Communication time breakdown.}
  Cumulative time by communication primitive for rank~0 of
  Qwen2.5-32B (32K tokens) under \texttt{dp1-tp8-pp4-cp1-RC}. The inset
  enlarges primitives with small shares.
  }
  \label{fig:qwen25_32b_comm_time_rank0}
\end{figure*}

\begin{figure*}[!htbp]
  \centering
  \includegraphics[width=0.96\textwidth]{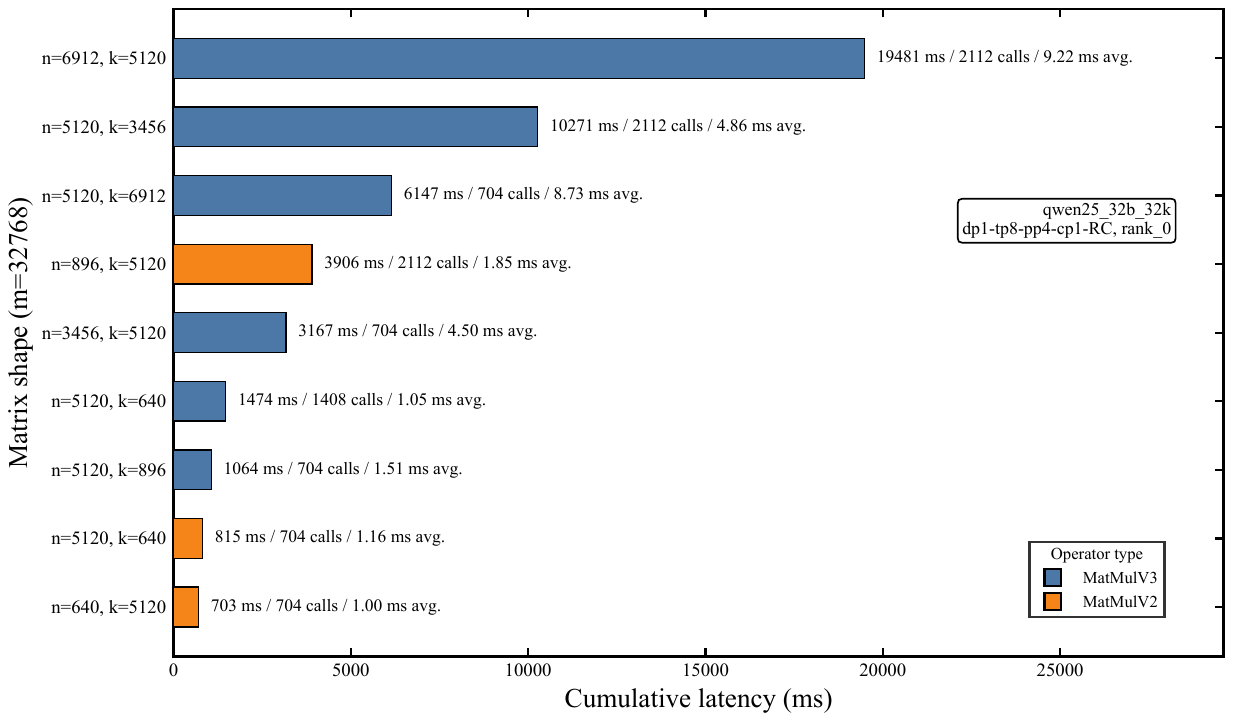}
  \caption{
  \textbf{Top matrix-operator shapes by cumulative latency.}
  Matrix operators on rank~0 are grouped by $(n,k)$ with $m=32768$ for
  Qwen2.5-32B (32K tokens) under \texttt{dp1-tp8-pp4-cp1-RC}. Each bar
  reports cumulative latency, call count, and mean latency.
  }
  \label{fig:qwen25_32b_matrix_operator_rank0}
\end{figure*}

\paragraph{Tracing dominant communication and matrix costs.}
The lower-level records identify the dominant costs. On rank~0 under
\texttt{dp1-tp8-pp4-cp1-RC}, \texttt{batchSendRecv} accounts for most
communication time (Figure~\ref{fig:qwen25_32b_comm_time_rank0}), while a
small set of matrix shapes accounts for most cumulative matrix-operator
latency (Figure~\ref{fig:qwen25_32b_matrix_operator_rank0}). These cumulative
records locate the primitives and shapes that account for device work;
the rank-level time breakdowns show how computation and communication
contribute to the execution schedule.

\paragraph{Relating operator estimates to iteration time.}
The Calculon baseline uses a compatibility setup with multi-head attention
in place of grouped-query attention and activation recomputation enabled;
the baseline implementation supports neither grouped-query attention nor
context parallelism. It predicts 73.76\,s per iteration against the measured
112.22\,s, a 34.3\% underestimate. The operator-level comparison in
Figure~\ref{fig:calculon-operator-latency-qwen25-32b-32k} shows that Calculon
recovers the dominant matrix shapes and principal \texttt{MatMulV3} hotspots,
but underestimates their cumulative cost by 27--35\%. Thus, the component
profiles expose underestimated operator costs alongside the end-to-end
runtime error.

\begin{figure*}[p]
\centering
\begin{subfigure}[t]{0.96\textwidth}
  \centering
  \includegraphics[width=\linewidth]{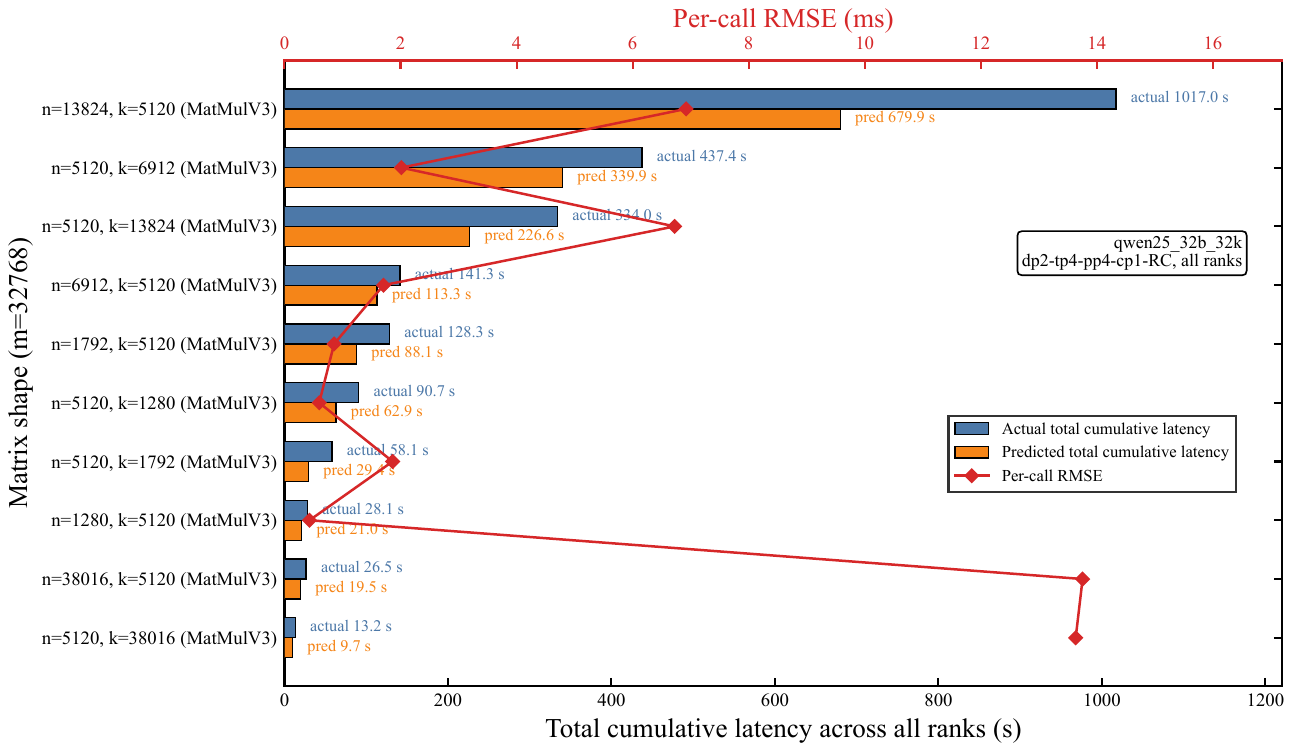}
  \caption{\texttt{dp2-tp4-pp4-cp1-RC}.}
  \label{fig:calculon-op-dp2-tp4-pp4-cp1-rc}
\end{subfigure}
\par\medskip
\begin{subfigure}[t]{0.96\textwidth}
  \centering
  \includegraphics[width=\linewidth]{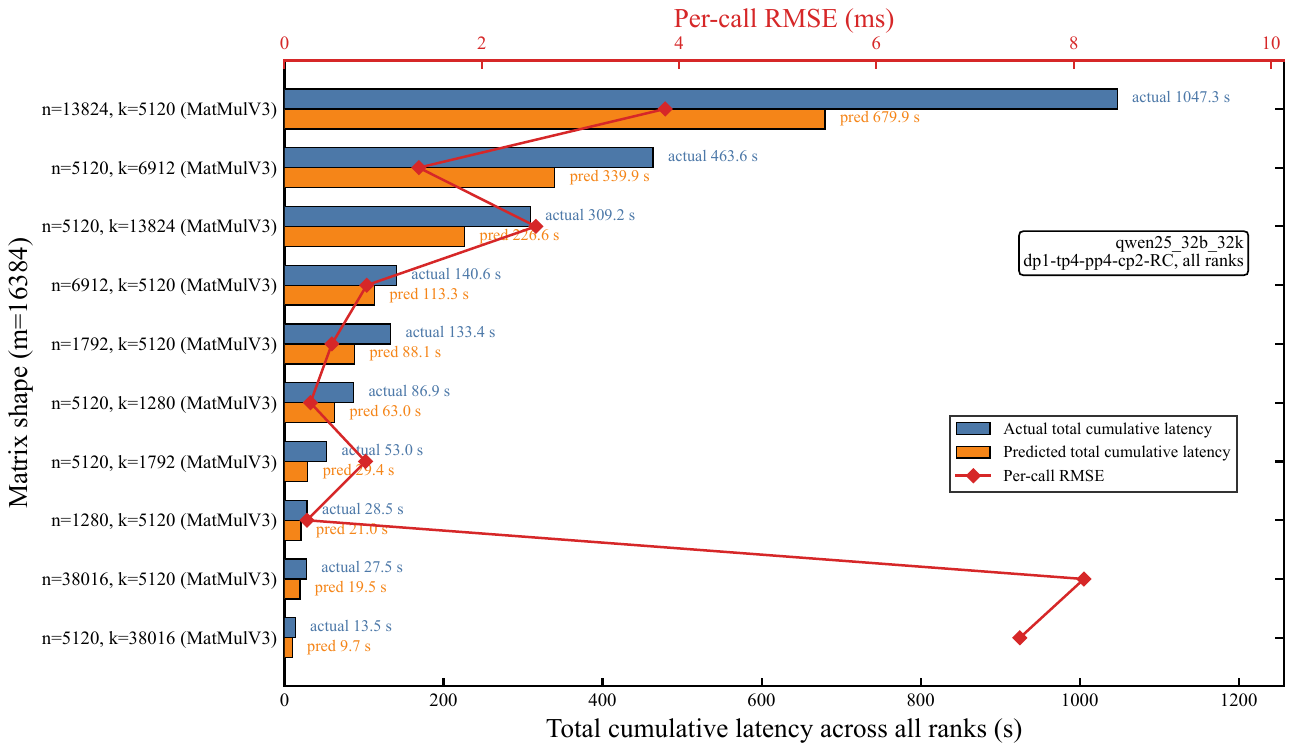}
  \caption{\texttt{dp1-tp4-pp4-cp2-RC}.}
  \label{fig:calculon-op-dp1-tp4-pp4-cp2-rc}
\end{subfigure}
\caption{
  Matrix-latency predictions for Qwen2.5-32B (32K tokens) under four
  parallel configurations. Bars compare measured and predicted cumulative
  latency across ranks for dominant shapes; the red curve shows per-call
  RMSE for each shape.
}
\label{fig:calculon-operator-latency-qwen25-32b-32k}
\end{figure*}

\begin{figure*}[p]
\ContinuedFloat
\centering
\begin{subfigure}[t]{0.96\textwidth}
  \centering
  \includegraphics[width=\linewidth]{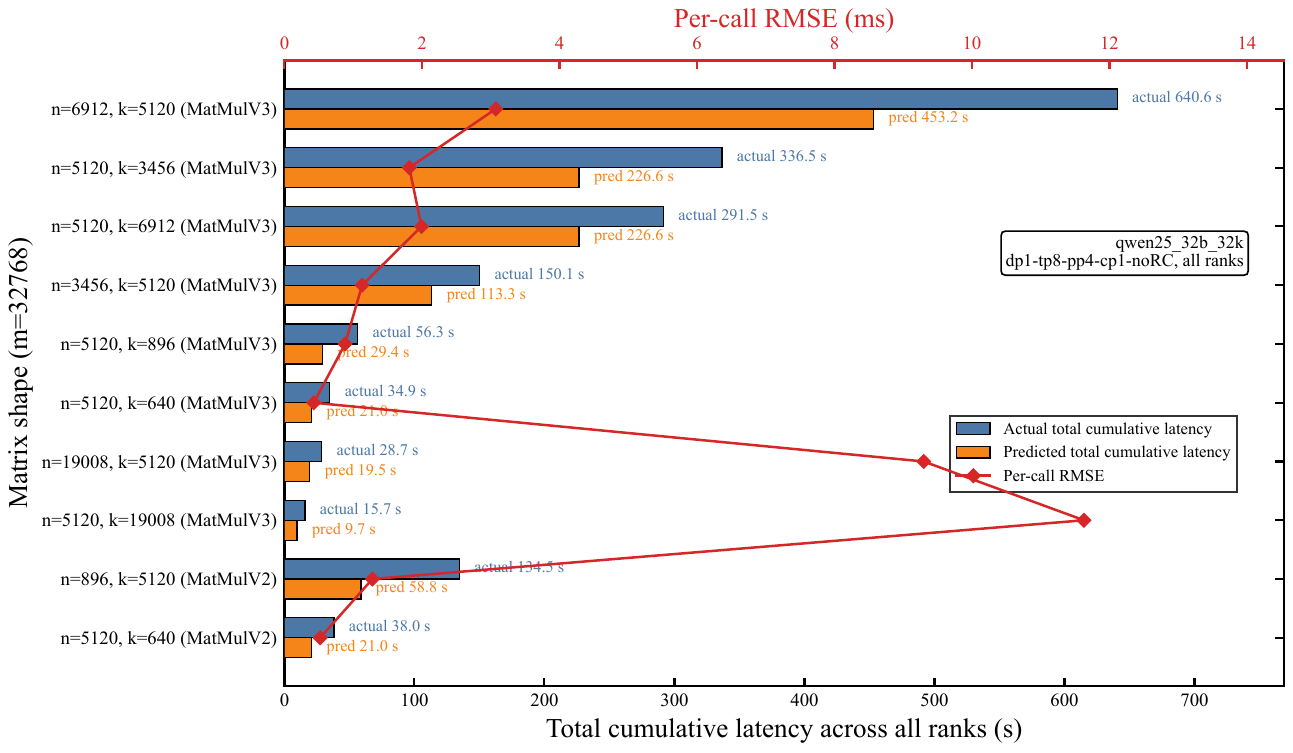}
  \caption{\texttt{dp1-tp8-pp4-cp1-noRC}.}
  \label{fig:calculon-op-dp1-tp8-pp4-cp1-norc}
\end{subfigure}
\par\medskip
\begin{subfigure}[t]{0.96\textwidth}
  \centering
  \includegraphics[width=\linewidth]{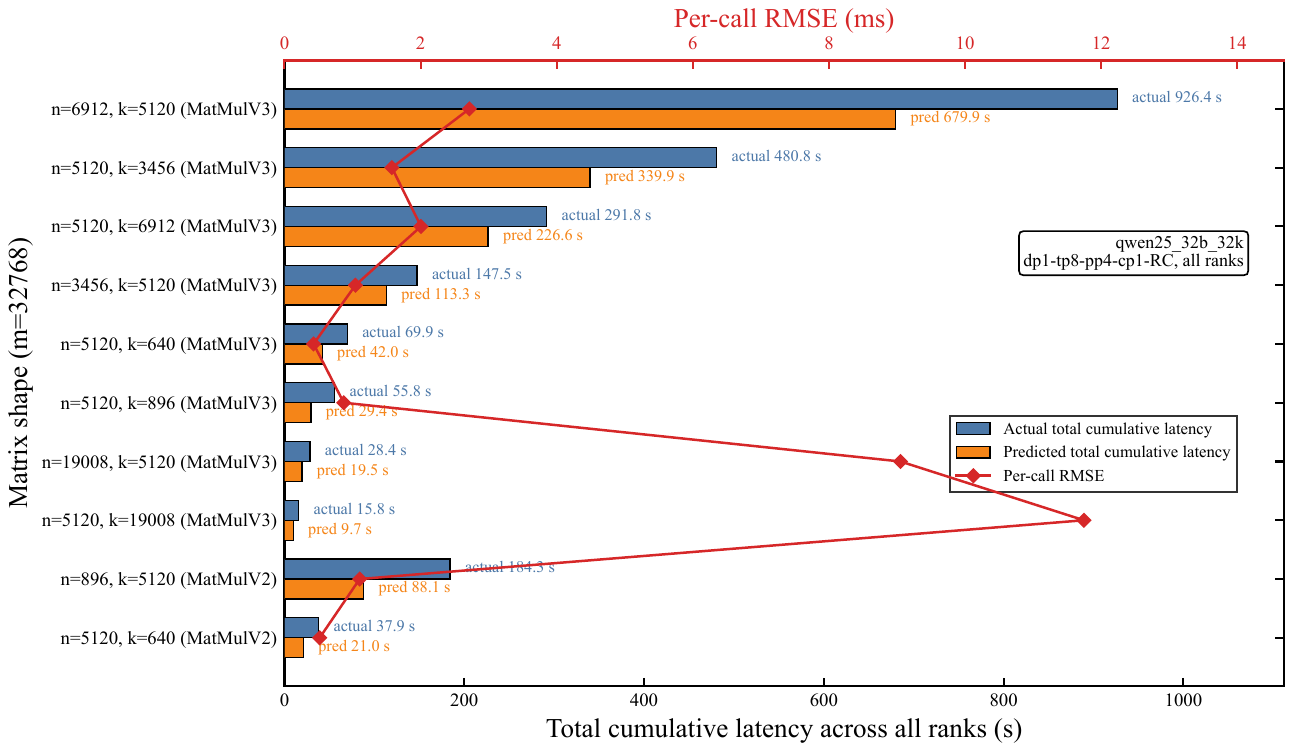}
  \caption{\texttt{dp1-tp8-pp4-cp1-RC}.}
  \label{fig:calculon-op-dp1-tp8-pp4-cp1-rc}
\end{subfigure}
\caption{
  Matrix-latency predictions for Qwen2.5-32B (32K tokens) under four
  parallel configurations. Bars compare measured and predicted cumulative
  latency across ranks for dominant shapes; the red curve shows per-call
  RMSE for each shape.
}

\end{figure*}

\FloatBarrier

\begin{table*}[p]
\begin{minipage}{\textwidth}
\subsection{Qwen1.5-1.8B Matrix-Operator Errors}
\label{app:detailed_matrix_errors}

Tables~\ref{tab:qwen15-matrix-metrics-by-config}
and~\ref{tab:qwen15-matrix-metrics-by-mnk} expand the aggregate matrix-operator
metrics in Table~\ref{tab:calculon-benchmark-summary} for Qwen1.5-1.8B
(8K tokens). Grouping by parallel configuration shows how estimation quality
varies with the training strategy; grouping by matrix dimensions identifies
the operator shapes associated with those errors. Together, the two tables
support examining a computation estimator at both configuration and kernel
granularity.
\end{minipage}
\par\medskip
\centering
\small
\caption{Matrix-operator prediction metrics by parallel configuration for Qwen1.5-1.8B (8K tokens). Configurations continue from the left panel to the right.}
\label{tab:qwen15-matrix-metrics-by-config}
\setlength{\tabcolsep}{3pt}
\begin{tabular*}{\textwidth}{@{\extracolsep{\fill}}lrrr@{\hspace{18pt}}lrrr@{}}
\toprule
Parallel configuration & MAE (ms) & sMAPE (\%) & $R^2$ & Parallel configuration & MAE (ms) & sMAPE (\%) & $R^2$ \\
\midrule
dp1-tp1-pp1-cp8-RC & 0.048 & 40.505 & 0.779 & dp2-tp1-pp1-cp4-RC & 0.072 & 36.826 & 0.864 \\
dp1-tp1-pp1-cp8-noRC & 0.053 & 42.190 & 0.778 & dp2-tp1-pp1-cp4-noRC & 0.076 & 38.311 & 0.873 \\
dp1-tp1-pp2-cp4-RC & 0.072 & 36.794 & 0.865 & dp2-tp1-pp2-cp2-RC & 0.120 & 30.403 & 0.890 \\
dp1-tp1-pp2-cp4-noRC & 0.076 & 38.126 & 0.874 & dp2-tp1-pp2-cp2-noRC & 0.127 & 31.581 & 0.897 \\
dp1-tp1-pp4-cp2-RC & 0.119 & 30.313 & 0.889 & dp2-tp1-pp4-cp1-RC & 0.252 & 31.580 & 0.874 \\
dp1-tp1-pp4-cp2-noRC & 0.128 & 31.596 & 0.895 & dp2-tp1-pp4-cp1-noRC & 0.268 & 32.662 & 0.880 \\
dp1-tp1-pp8-cp1-RC & 0.251 & 31.436 & 0.875 & dp2-tp2-pp1-cp2-RC & 0.101 & 45.635 & 0.758 \\
dp1-tp1-pp8-cp1-noRC & 0.267 & 32.491 & 0.881 & dp2-tp2-pp1-cp2-noRC & 0.118 & 50.418 & 0.748 \\
dp1-tp2-pp1-cp4-RC & 0.056 & 42.201 & 0.725 & dp2-tp2-pp2-cp1-RC & 0.203 & 42.782 & 0.741 \\
dp1-tp2-pp1-cp4-noRC & 0.061 & 45.811 & 0.741 & dp2-tp2-pp2-cp1-noRC & 0.245 & 48.749 & 0.722 \\
dp1-tp2-pp2-cp2-RC & 0.093 & 43.991 & 0.792 & dp2-tp4-pp1-cp1-RC & 0.117 & 43.081 & 0.696 \\
dp1-tp2-pp2-cp2-noRC & 0.113 & 49.147 & 0.769 & dp2-tp4-pp1-cp1-noRC & 0.141 & 47.530 & 0.690 \\
dp1-tp2-pp4-cp1-RC & 0.203 & 42.833 & 0.741 & dp4-tp1-pp1-cp2-RC & 0.120 & 30.453 & 0.888 \\
dp1-tp2-pp4-cp1-noRC & 0.245 & 48.783 & 0.721 & dp4-tp1-pp1-cp2-noRC & 0.127 & 31.512 & 0.894 \\
dp1-tp4-pp1-cp2-RC & 0.060 & 39.834 & 0.731 & dp4-tp1-pp2-cp1-RC & 0.262 & 32.662 & 0.869 \\
dp1-tp4-pp1-cp2-noRC & 0.069 & 43.331 & 0.731 & dp4-tp1-pp2-cp1-noRC & 0.271 & 32.963 & 0.879 \\
dp1-tp4-pp2-cp1-RC & 0.112 & 41.730 & 0.700 & dp4-tp2-pp1-cp1-RC & 0.210 & 43.702 & 0.727 \\
dp1-tp4-pp2-cp1-noRC & 0.144 & 48.204 & 0.690 & dp4-tp2-pp1-cp1-noRC & 0.245 & 48.707 & 0.721 \\
dp1-tp8-pp1-cp1-RC & 0.107 & 60.527 & 0.608 & dp8-tp1-pp1-cp1-RC & 0.279 & 34.472 & 0.859 \\
dp1-tp8-pp1-cp1-noRC & 0.109 & 61.202 & 0.655 & dp8-tp1-pp1-cp1-noRC & 0.292 & 34.973 & 0.859 \\
\bottomrule
\end{tabular*}
\end{table*}

\begin{table*}[p]
\centering
\small
\caption{Matrix-operator prediction metrics by matrix dimensions for Qwen1.5-1.8B (8K tokens). Shapes continue from the left panel to the right.}
\label{tab:qwen15-matrix-metrics-by-mnk}
\setlength{\tabcolsep}{3pt}
\begin{tabular*}{\textwidth}{@{\extracolsep{\fill}}rrrrr@{\hspace{18pt}}rrrrr@{}}
\toprule
$m$ & $n$ & $k$ & MAE (ms) & sMAPE (\%) & $m$ & $n$ & $k$ & MAE (ms) & sMAPE (\%) \\
\midrule
1024 & 2048 & 2048 & 0.011 & 24.151 & 4096 & 3072 & 2048 & 0.078 & 38.374 \\
1024 & 2048 & 5504 & 0.056 & 55.184 & 4096 & 5504 & 2048 & 0.114 & 32.488 \\
1024 & 2048 & 6144 & 0.068 & 58.875 & 4096 & 6144 & 2048 & 0.108 & 28.405 \\
1024 & 2048 & 11008 & 0.089 & 46.131 & 4096 & 11008 & 2048 & 0.193 & 29.637 \\
1024 & 5504 & 2048 & 0.046 & 47.035 & 4096 & 37984 & 2048 & 1.271 & 48.828 \\
1024 & 6144 & 2048 & 0.048 & 45.422 & 4096 & 75968 & 2048 & 1.350 & 30.161 \\
1024 & 11008 & 2048 & 0.064 & 35.649 & 4096 & 151936 & 2048 & 2.326 & 26.538 \\
1024 & 151936 & 2048 & 0.968 & 40.604 & 8192 & 256 & 2048 & 0.018 & 36.910 \\
2048 & 1024 & 2048 & 0.009 & 19.450 & 8192 & 512 & 2048 & 0.032 & 45.419 \\
2048 & 2048 & 1024 & 0.002 & 4.102 & 8192 & 688 & 2048 & 0.191 & 113.572 \\
2048 & 2048 & 2048 & 0.031 & 44.121 & 8192 & 768 & 2048 & 0.078 & 64.092 \\
2048 & 2048 & 2752 & 0.077 & 68.929 & 8192 & 1024 & 2048 & 0.042 & 32.793 \\
2048 & 2048 & 3072 & 0.082 & 63.776 & 8192 & 1376 & 2048 & 0.091 & 47.508 \\
2048 & 2048 & 5504 & 0.096 & 48.069 & 8192 & 1536 & 2048 & 0.078 & 38.603 \\
2048 & 2048 & 6144 & 0.075 & 37.723 & 8192 & 2048 & 256 & 0.009 & 20.407 \\
2048 & 2048 & 11008 & 0.153 & 41.648 & 8192 & 2048 & 512 & 0.030 & 43.394 \\
2048 & 2752 & 2048 & 0.062 & 59.668 & 8192 & 2048 & 688 & 0.234 & 120.810 \\
2048 & 3072 & 2048 & 0.050 & 47.278 & 8192 & 2048 & 768 & 0.032 & 33.151 \\
2048 & 5504 & 2048 & 0.070 & 38.430 & 8192 & 2048 & 1024 & 0.062 & 42.996 \\
2048 & 6144 & 2048 & 0.073 & 36.576 & 8192 & 2048 & 1376 & 0.092 & 47.680 \\
2048 & 11008 & 2048 & 0.093 & 27.610 & 8192 & 2048 & 1536 & 0.121 & 48.883 \\
2048 & 75968 & 2048 & 0.736 & 32.456 & 8192 & 2048 & 2048 & 0.074 & 28.819 \\
2048 & 151936 & 2048 & 1.269 & 28.616 & 8192 & 2048 & 2752 & 0.177 & 45.856 \\
4096 & 512 & 2048 & 0.009 & 20.546 & 8192 & 2048 & 3072 & 0.545 & 88.956 \\
4096 & 1024 & 2048 & 0.035 & 48.722 & 8192 & 2048 & 5504 & 0.435 & 54.585 \\
4096 & 1376 & 2048 & 0.066 & 62.631 & 8192 & 2048 & 6144 & 0.189 & 26.503 \\
4096 & 1536 & 2048 & 0.047 & 44.944 & 8192 & 2048 & 11008 & 0.387 & 29.873 \\
4096 & 2048 & 512 & 0.005 & 12.478 & 8192 & 2048 & 18992 & 1.663 & 60.863 \\
4096 & 2048 & 1024 & 0.039 & 51.339 & 8192 & 2048 & 37984 & 5.344 & 82.560 \\
4096 & 2048 & 1376 & 0.065 & 61.595 & 8192 & 2752 & 2048 & 0.139 & 38.224 \\
4096 & 2048 & 1536 & 0.038 & 37.719 & 8192 & 3072 & 2048 & 0.137 & 34.575 \\
4096 & 2048 & 2048 & 0.044 & 33.438 & 8192 & 5504 & 2048 & 0.268 & 38.426 \\
4096 & 2048 & 2752 & 0.099 & 50.468 & 8192 & 6144 & 2048 & 0.259 & 34.041 \\
4096 & 2048 & 3072 & 0.192 & 68.259 & 8192 & 11008 & 2048 & 0.440 & 32.862 \\
4096 & 2048 & 5504 & 0.170 & 43.264 & 8192 & 18992 & 2048 & 1.423 & 53.131 \\
4096 & 2048 & 6144 & 0.079 & 21.620 & 8192 & 37984 & 2048 & 2.795 & 52.109 \\
4096 & 2048 & 11008 & 0.263 & 38.596 & 8192 & 75968 & 2048 & 2.778 & 30.899 \\
4096 & 2048 & 37984 & 2.144 & 72.136 & 8192 & 151936 & 2048 & 5.433 & 30.293 \\
4096 & 2752 & 2048 & 0.081 & 43.180 &  &  &  &  &  \\
\bottomrule
\end{tabular*}
\end{table*}

\FloatBarrier
\section{Cross-Platform Evaluation}
\label{app:cross_platform}

This section expands the platform-calibration and configuration-selection
results in Sections~\ref{subsec:platform-calibration}
and~\ref{subsec:cross-platform-results}. Paired V100/910B3 profiles support
two uses of LM-SimBench: calibrating a common simulator on each platform and
using NPU measurements to select promising GPU configurations. Detailed
time, memory, and component records connect these uses to the execution
characteristics of the two stacks.

\subsection{Experimental Setup and Coverage}
\label{app:cross_platform_scope}
\label{app:cross_platform_coverage}

We collect new paired measurements on 8$\times$V100 (32\,GiB per device) and
8$\times$910B3 (64\,GiB) for Qwen3-0.6B and Qwen3-1.7B at a training
sequence length of 4K tokens. Each model uses the same 18 DP/TP/PP/recomputation
candidates on both stacks, matching the CP${=}1$ subset of its 38
configurations in the primary release. The NPU reference in this experiment
was collected separately from that release. Both stacks use a distributed
optimizer and profile the sixth iteration after five warm-up steps.
The supplementary data directories use the suffix
\texttt{\_tp\_pp\_sweep\_cp1\_npu} for NPU runs and
\texttt{\_tp\_pp\_sweep\_cp1\_gpu} under \texttt{processed\_gpu/} for GPU
runs, prefixed by the workload name.

Table~\ref{tab:gpu-alignment} lists the matched workload settings and the
platform-specific precision, fusion, sequence-parallel, and profiler settings.
Ascend-specific \texttt{AICore}/\texttt{MTE}/\texttt{cube} and
\texttt{HCCS}/\texttt{RDMA}/\texttt{SDMA} fields are left empty in GPU records.
Driver, CUDA/CANN, PyTorch, collective-library, and framework revisions are
recorded in the artifact's environment metadata.

\begin{table*}[t]
\centering
\small
\caption{Matched workload settings and platform-specific choices in the
cross-platform experiment.}
\label{tab:gpu-alignment}
\begin{tabularx}{\linewidth}{>{\RaggedRight\arraybackslash}p{0.18\linewidth}
                              >{\RaggedRight\arraybackslash}X}
\toprule
\textbf{Aspect} & \textbf{Setting} \\
\midrule
Matched
& Model architecture, layer counts, hidden size, attention heads, and
grouped-query attention; sequence length 4096; micro/global batch 1/32;
eight devices; CP${=}1$; candidate configuration names; optimizer
hyperparameters and seed; profiled iteration index. \\
\addlinespace[2pt]
Stack-specific
& Data type (FP16 on V100 vs.\ BF16 on 910B3); fused kernels (FlashAttention
and fused RoPE/SwiGLU/RMSNorm disabled in the GPU sweep, enabled on the Ascend
stack); sequence parallelism (enabled on the NPU side); profiler-visible
hardware counters and communication-link fields; input data path. \\
\bottomrule
\end{tabularx}
\end{table*}

Complete GPU profiles are available for 15 of 18 candidates for
Qwen3-0.6B and 12 of 18 for Qwen3-1.7B; the NPU
side is complete for 18 and 16 candidates, respectively. Paired statistics
therefore use 15 and 11 same-named configurations with complete profiles on
both stacks, 26 pairs in total. Table~\ref{tab:gpu-coverage} lists every
candidate that is not complete on both stacks. Empty GPU profile directories
are marked \texttt{missing\_or\_incomplete} in the manifest; their failure
causes are unknown. Table~\ref{tab:cross-platform-summary} summarizes the
paired comparisons.

\begin{table*}[t]
\centering
\footnotesize
\caption{Candidates with missing or incomplete profiles on either stack,
excluded from paired comparisons.}
\label{tab:gpu-coverage}
\begin{tabularx}{\linewidth}{@{}p{0.14\linewidth}p{0.30\linewidth}p{0.38\linewidth}X@{}}
\toprule
Model & Configuration & GPU status & NPU status \\
\midrule
Qwen3-0.6B & \texttt{dp2-tp1-pp4-cp1-noRC} & Missing/incomplete (empty directory) & Complete \\
Qwen3-0.6B & \texttt{dp4-tp1-pp2-cp1-noRC} & Missing/incomplete (empty directory) & Complete \\
Qwen3-0.6B & \texttt{dp8-tp1-pp1-cp1-noRC} & Missing/incomplete (empty directory) & Complete \\
Qwen3-1.7B & \texttt{dp1-tp2-pp4-cp1-noRC} & Missing/incomplete (empty directory) & Complete \\
Qwen3-1.7B & \texttt{dp2-tp1-pp4-cp1-noRC} & Missing/incomplete (empty directory) & Complete \\
Qwen3-1.7B & \texttt{dp2-tp2-pp2-cp1-noRC} & Missing/incomplete (empty directory) & Complete \\
Qwen3-1.7B & \texttt{dp4-tp1-pp2-cp1-RC} & Complete & Missing \\
Qwen3-1.7B & \texttt{dp4-tp1-pp2-cp1-noRC} & Missing/incomplete (empty directory) & Missing \\
Qwen3-1.7B & \texttt{dp4-tp2-pp1-cp1-noRC} & Missing/incomplete (empty directory) & Complete \\
Qwen3-1.7B & \texttt{dp8-tp1-pp1-cp1-noRC} & Missing/incomplete (empty directory) & Complete \\
\bottomrule
\end{tabularx}
\end{table*}

\begin{table*}[t]
\centering
\small
\caption{Cross-platform agreement in configuration rankings, memory behavior,
and operator performance between V100 GPUs and Ascend 910B3 NPUs.
$^\ast$Three paired configurations.}
\label{tab:cross-platform-summary}
\begin{tabular}{llcc}
\toprule
Level & Metric & Qwen3-0.6B & Qwen3-1.7B \\
\midrule
Coverage & Paired complete configurations & 15 & 11 \\
\midrule
Strategy & Iteration rank $\rho$ / $\tau_b$, pooled & 0.718 / 0.562 & 0.664 / 0.491 \\
         & Iteration rank $\rho$, noRC / RC & 0.714 / 0.683 & 1.000$^\ast$ / 0.690 \\
\midrule
System   & Peak-memory rank $\rho$ / $\tau_b$ & 0.982 / 0.924 & 0.991 / 0.964 \\
         & Lowest-memory Top-3 overlap & 3/3 & 3/3 \\
         & RC time/memory direction agreement & 6 / 6 & 3 / 3 \\
\midrule
Component & Matrix median-latency $\rho$ / $\tau_b$ & 0.983 / 0.892 & 0.987 / 0.909 \\
          & Slowest matrix-problem Top-10 overlap & 8/10 & 10/10 \\
          & Operator-category median $\rho$ / JSD & 0.503 / 0.436 & 0.531 / 0.356 \\
          & Communication-primitive median $\rho$ / JSD & 0.949 / 0.019 & 0.872 / 0.035 \\
          & Non-overlapped comm.\ share, GPU / NPU & 17.0\% / 28.5\% & 10.9\% / 30.0\% \\
\bottomrule
\end{tabular}
\end{table*}

\subsection{Simulator Calibration across Platforms}
\label{app:cross_platform_calculon}

Calculon provides a concrete application of the profiling data to simulator
calibration and evaluation. We use a common modeling implementation on both
platforms, calibrate its platform parameters against the corresponding
profiles, and compare its time, memory, and ranking outputs with the measured
targets. The implementation includes Qwen3 architecture and memory accounting,
kernel launch latency, pipeline scheduling, and rank-level memory output.
Each platform uses one parameter set for compute efficiency, launch latency,
communication, and runtime memory across both models and the full candidate
grid. Simulation outputs are stored under
\texttt{calculon/outputs/qwen3\_sweeps\_final/}.

Table~\ref{tab:gpu-calculon} expands Table~\ref{tab:main-calibration} with
rank-level memory errors. On the platform-calibration configurations,
iteration-time MAPE is
$7.97\%$--$11.22\%$ and
time-ranking $\rho$ is $0.853$--$0.950$; rank-level peak-memory MAPE is
$12.55\%$--$15.99\%$ and configuration-level peak-memory ranking $\rho$ is
$0.986$--$0.994$. Across the four model$\times$recomputation settings, GPU
measurement, NPU measurement, Calculon-GPU, and Calculon-NPU select the same
Top-1 configuration (4/4).

\begin{table*}[t]
\centering
\small
\caption{Calculon time, memory, and ranking accuracy on the configurations
used for platform-specific calibration.}
\label{tab:gpu-calculon}
\begin{tabular}{llrrrr}
\toprule
Model & Platform & Time MAPE (\%) & Time $\rho$ & Rank-mem.\ MAPE (\%) & Config-mem.\ $\rho$ \\
\midrule
Qwen3-0.6B & V100 & 10.28 & 0.861 & 15.99 & 0.993 \\
Qwen3-0.6B & 910B3 & 10.19 & 0.950 & 13.20 & 0.986 \\
Qwen3-1.7B & V100 & 11.22 & 0.853 & 12.56 & 0.993 \\
Qwen3-1.7B & 910B3 & 7.97 & 0.935 & 12.55 & 0.994 \\
\bottomrule
\end{tabular}
\end{table*}

\subsection{NPU-Guided GPU Configuration Selection}
\label{app:cross_platform_strategy}

NPU measurements provide a shortlist for GPU configuration selection,
following the procedure in Section~\ref{subsec:cross-platform-metrics}.
For each model, we retrospectively rank the paired-complete candidates by measured NPU
iteration time, retain the fastest three, and identify the fastest GPU
configuration within that shortlist using its measured GPU time. The
comparison target is the fastest GPU configuration over the entire paired
candidate set. Top-1 selection regret and Top-3 shortlist regret measure the GPU runtime
penalty relative to this optimum when selecting the NPU-fastest candidate or
the GPU-fastest member of the NPU-derived top-three shortlist, respectively
(Eq.~\ref{eq:benchmark-shortlist-regret}).
The values are recomputed from the rounded times in Table~\ref{tab:gpu-time}.
In both workloads the NPU-derived shortlist contains this
GPU optimum, giving zero shortlist selection regret
(Table~\ref{tab:gpu-shortlist}). The candidate counts decrease from 15 to 3
and from 11 to 3, respectively.

\begin{table*}[t]
\centering
\small
\caption{GPU configuration selection using NPU-derived rankings and top-three shortlists.}
\label{tab:gpu-shortlist}
\begin{tabular}{lrrrr}
\toprule
Model & Paired candidates & Top-1 regret & GPU optimum in Top-3 & Top-3 regret \\
\midrule
Qwen3-0.6B & 15 & 0.0\% & Yes & 0.0\% \\
Qwen3-1.7B & 11 & 6.9\% & Yes & 0.0\% \\
\bottomrule
\end{tabular}
\end{table*}

Pooled over both recomputation settings, iteration-time rank agreement between
the two stacks is moderate ($\rho=0.718$ and $0.664$, $\tau_b=0.562$ and
$0.491$ for the two models). Directly choosing the NPU-fastest configuration
finds the GPU optimum for Qwen3-0.6B and incurs a 6.9\% GPU runtime penalty
for Qwen3-1.7B. Table~\ref{tab:gpu-strata} additionally compares configurations
with recomputation fixed: within each of the four
model$\times$recomputation strata, the two stacks select the same Top-1
configuration. This conditional agreement and the pooled shortlist results
show how NPU observations can narrow GPU search even when the complete
rankings differ.

\begin{table*}[t]
\centering
\small
\caption{Cross-platform iteration-time rank agreement by recomputation setting.}
\label{tab:gpu-strata}
\begin{tabular}{llrrrrc}
\toprule
Model & Stratum & $n$ & $\rho$ & $\tau_b$ & Top-3 & Top-1 agreement \\
\midrule
Qwen3-0.6B & all & 15 & 0.718 & 0.562 & 2/3 & yes \\
Qwen3-0.6B & noRC & 6 & 0.714 & 0.600 & 2/3 & yes \\
Qwen3-0.6B & RC & 9 & 0.683 & 0.556 & 2/3 & yes \\
\midrule
Qwen3-1.7B & all & 11 & 0.664 & 0.491 & 2/3 & \textbf{no} \\
Qwen3-1.7B & noRC & 3 & 1.000 & 1.000 & 3/3 & yes \\
Qwen3-1.7B & RC & 8 & 0.690 & 0.571 & 2/3 & yes \\
\bottomrule
\end{tabular}
\end{table*}

Table~\ref{tab:gpu-time} reports the per-configuration iteration times, with
ranks computed over all paired configurations within each model, pooling RC
and noRC. A leave-one-configuration-out diagnostic fits a median GPU/NPU time ratio on
the other paired configurations and applies it to the held-out NPU time.
It yields $22.4\%$ and $22.9\%$ sMAPE with $R^2=-8.33$ and $-4.74$,
respectively, showing that the runtime difference varies with configuration.

\begin{table*}[t]
\centering
\footnotesize
\caption{GPU and NPU iteration times by configuration, ordered by GPU runtime within each model.}
\label{tab:gpu-time}
\begin{tabular}{lrrrrr}
\toprule
Configuration & GPU (s) & NPU (s) & GPU/NPU & GPU rank & NPU rank \\
\midrule
\multicolumn{6}{@{}l}{\textit{Qwen3-0.6B}} \\
\texttt{dp4-tp2-pp1-cp1-noRC} & 6.40 & 3.93 & 1.63 & 1 & 1 \\
\texttt{dp2-tp4-pp1-cp1-noRC} & 6.69 & 6.11 & 1.09 & 2 & 7 \\
\texttt{dp2-tp2-pp2-cp1-noRC} & 7.20 & 4.48 & 1.61 & 3 & 2 \\
\texttt{dp1-tp4-pp2-cp1-noRC} & 7.33 & 6.74 & 1.09 & 4 & 10 \\
\texttt{dp8-tp1-pp1-cp1-RC} & 7.53 & 4.69 & 1.61 & 5 & 3 \\
\texttt{dp1-tp2-pp4-cp1-noRC} & 8.28 & 5.10 & 1.62 & 6 & 4 \\
\texttt{dp4-tp2-pp1-cp1-RC} & 8.60 & 5.63 & 1.53 & 7 & 6 \\
\texttt{dp2-tp4-pp1-cp1-RC} & 8.93 & 9.48 & 0.94 & 8 & 12 \\
\texttt{dp4-tp1-pp2-cp1-RC} & 8.95 & 5.50 & 1.63 & 9 & 5 \\
\texttt{dp1-tp8-pp1-cp1-noRC} & 9.06 & 11.39 & 0.80 & 10 & 14 \\
\texttt{dp2-tp2-pp2-cp1-RC} & 9.46 & 6.27 & 1.51 & 11 & 9 \\
\texttt{dp1-tp4-pp2-cp1-RC} & 9.69 & 9.75 & 0.99 & 12 & 13 \\
\texttt{dp2-tp1-pp4-cp1-RC} & 10.25 & 6.25 & 1.64 & 13 & 8 \\
\texttt{dp1-tp2-pp4-cp1-RC} & 10.65 & 6.98 & 1.53 & 14 & 11 \\
\texttt{dp1-tp8-pp1-cp1-RC} & 12.86 & 18.36 & 0.70 & 15 & 15 \\
\midrule
\multicolumn{6}{@{}l}{\textit{Qwen3-1.7B}} \\
\texttt{dp2-tp4-pp1-cp1-noRC} & 8.85 & 6.42 & 1.38 & 1 & 2 \\
\texttt{dp8-tp1-pp1-cp1-RC} & 9.46 & 5.49 & 1.73 & 2 & 1 \\
\texttt{dp1-tp4-pp2-cp1-noRC} & 9.82 & 7.43 & 1.32 & 3 & 4 \\
\texttt{dp1-tp8-pp1-cp1-noRC} & 10.75 & 12.09 & 0.89 & 4 & 10 \\
\texttt{dp4-tp2-pp1-cp1-RC} & 11.20 & 7.04 & 1.59 & 5 & 3 \\
\texttt{dp2-tp4-pp1-cp1-RC} & 11.74 & 9.26 & 1.27 & 6 & 8 \\
\texttt{dp2-tp2-pp2-cp1-RC} & 12.34 & 7.83 & 1.58 & 7 & 6 \\
\texttt{dp1-tp4-pp2-cp1-RC} & 12.82 & 10.14 & 1.26 & 8 & 9 \\
\texttt{dp2-tp1-pp4-cp1-RC} & 13.28 & 7.46 & 1.78 & 9 & 5 \\
\texttt{dp1-tp2-pp4-cp1-RC} & 14.10 & 8.69 & 1.62 & 10 & 7 \\
\texttt{dp1-tp8-pp1-cp1-RC} & 14.28 & 16.78 & 0.85 & 11 & 11 \\
\bottomrule
\end{tabular}
\end{table*}

\subsection{Memory and Recomputation}
\label{app:cross_platform_memory}

Configuration-level peak memory is the maximum across ranks, matching the
memory-ranking task in Section~\ref{sec:benchmark}. Its ordering is strongly
aligned across platforms: rank correlation is $\rho=0.982$ and
$0.991$ ($\tau_b=0.924$ and $0.964$), and the three lowest-memory
configurations coincide on both stacks for both models. The median GPU/NPU
memory ratios are 0.77 and 0.70. Table~\ref{tab:gpu-memory} lists the
per-configuration peak memory.

This agreement also holds when recomputation is fixed: the RC/noRC
peak-memory rank correlations are 0.983/0.943 for Qwen3-0.6B and
0.976/1.000 for Qwen3-1.7B (the latter noRC subset contains three candidates).
The agreement across these subsets supports using NPU profiles to identify
relatively memory-efficient GPU strategies under either recomputation
setting.

\begin{table*}[t]
\centering
\footnotesize
\caption{GPU and NPU peak memory by configuration, measured as the maximum across ranks.}
\label{tab:gpu-memory}
\begin{tabular}{lrrrrr}
\toprule
Configuration & GPU (GiB) & NPU (GiB) & GPU/NPU & GPU rank & NPU rank \\
\midrule
\multicolumn{6}{@{}l}{\textit{Qwen3-0.6B}} \\
\texttt{dp1-tp8-pp1-cp1-RC} & 1.47 & 2.18 & 0.67 & 1 & 1 \\
\texttt{dp1-tp4-pp2-cp1-RC} & 2.18 & 3.16 & 0.69 & 2 & 2 \\
\texttt{dp2-tp4-pp1-cp1-RC} & 2.37 & 3.25 & 0.73 & 3 & 3 \\
\texttt{dp1-tp2-pp4-cp1-RC} & 3.64 & 5.26 & 0.69 & 4 & 5 \\
\texttt{dp2-tp2-pp2-cp1-RC} & 3.67 & 5.09 & 0.72 & 5 & 4 \\
\texttt{dp4-tp2-pp1-cp1-RC} & 4.14 & 5.37 & 0.77 & 6 & 6 \\
\texttt{dp2-tp1-pp4-cp1-RC} & 6.44 & 8.91 & 0.72 & 7 & 7 \\
\texttt{dp4-tp1-pp2-cp1-RC} & 6.68 & 8.94 & 0.75 & 8 & 9 \\
\texttt{dp8-tp1-pp1-cp1-RC} & 7.70 & 9.63 & 0.80 & 9 & 10 \\
\texttt{dp1-tp8-pp1-cp1-noRC} & 8.55 & 8.92 & 0.96 & 10 & 8 \\
\texttt{dp1-tp4-pp2-cp1-noRC} & 14.99 & 13.23 & 1.13 & 11 & 11 \\
\texttt{dp2-tp4-pp1-cp1-noRC} & 15.41 & 14.25 & 1.08 & 12 & 12 \\
\texttt{dp1-tp2-pp4-cp1-noRC} & 28.11 & 22.97 & 1.22 & 13 & 14 \\
\texttt{dp2-tp2-pp2-cp1-noRC} & 28.13 & 22.74 & 1.24 & 14 & 13 \\
\texttt{dp4-tp2-pp1-cp1-noRC} & 29.03 & 24.91 & 1.17 & 15 & 15 \\
\midrule
\multicolumn{6}{@{}l}{\textit{Qwen3-1.7B}} \\
\texttt{dp1-tp8-pp1-cp1-RC} & 2.81 & 4.54 & 0.62 & 1 & 1 \\
\texttt{dp1-tp4-pp2-cp1-RC} & 3.61 & 5.72 & 0.63 & 2 & 2 \\
\texttt{dp2-tp4-pp1-cp1-RC} & 4.26 & 6.13 & 0.69 & 3 & 3 \\
\texttt{dp1-tp2-pp4-cp1-RC} & 5.38 & 8.34 & 0.64 & 4 & 5 \\
\texttt{dp2-tp2-pp2-cp1-RC} & 5.71 & 8.24 & 0.69 & 5 & 4 \\
\texttt{dp4-tp2-pp1-cp1-RC} & 7.16 & 9.31 & 0.77 & 6 & 6 \\
\texttt{dp2-tp1-pp4-cp1-RC} & 8.92 & 12.73 & 0.70 & 7 & 7 \\
\texttt{dp1-tp8-pp1-cp1-noRC} & 11.24 & 14.02 & 0.80 & 8 & 8 \\
\texttt{dp8-tp1-pp1-cp1-RC} & 12.96 & 15.66 & 0.83 & 9 & 9 \\
\texttt{dp1-tp4-pp2-cp1-noRC} & 18.10 & 18.85 & 0.96 & 10 & 10 \\
\texttt{dp2-tp4-pp1-cp1-noRC} & 18.96 & 20.20 & 0.94 & 11 & 11 \\
\bottomrule
\end{tabular}
\end{table*}

Recomputation behaves consistently across stacks. Across six matched base
configurations for Qwen3-0.6B and three for
Qwen3-1.7B, enabling recomputation increases iteration time and
decreases peak memory in the same direction in every pair (100\% directional
agreement). The median time overhead is 32.9\% (GPU) versus 44.0\% (NPU) for
the smaller model and 32.6\% versus 38.8\% for the larger one; the median
memory saving is 85.6\% versus 77.2\% and 77.5\% versus 69.7\%, respectively.

\subsection{Operator and Communication Behavior}
\label{app:cross_platform_component}

\paragraph{Matrix-problem latency.}
Matrix operators are matched by batch size, dimensions $(m,n,k)$, and
recorded input-data-type labels. Type labels are normalized for capitalization,
quotation marks, and whitespace before matching.
Per-call latencies are aggregated across all paired configurations and all
eight ranks to obtain a median for each problem on each stack. Each
model has 27 shared problem types, and those types account for 72.8\% and
76.7\% of GPU matrix calls. Ranking the shared problems by median latency gives
$\rho=0.983$ and $0.987$ ($\tau_b=0.892$ and $0.909$), with slowest-shape
Top-10 overlap of $8/10$ and $10/10$. The relative latency ordering of the
shared matrix problems is therefore preserved, whereas absolute speed differs
by a median factor of 2.79 and 2.84 (Table~\ref{tab:gpu-matrix}).

\begin{table*}[t]
\centering
\footnotesize
\caption{Median per-call latency of representative matched matrix operators on GPUs and NPUs.}
\label{tab:gpu-matrix}
\begin{tabular}{llrrr}
\toprule
Model & Matrix problem & GPU (\textmu s) & NPU (\textmu s) & GPU/NPU \\
\midrule
Qwen3-0.6B & \texttt{b8: 4096x4096 @ 4096x128} & 654.5 & 388.4 & 1.69 \\
Qwen3-0.6B & \texttt{b8: 4096x128 @ 128x4096} & 693.3 & 234.0 & 2.96 \\
Qwen3-0.6B & \texttt{b16: 4096x4096 @ 4096x128} & 1187.3 & 697.2 & 1.70 \\
Qwen3-0.6B & \texttt{b4: 4096x4096 @ 4096x128} & 278.4 & 133.5 & 2.09 \\
Qwen3-0.6B & \texttt{b4: 4096x128 @ 128x4096} & 352.4 & 97.0 & 3.63 \\
Qwen3-0.6B & \texttt{b16: 4096x128 @ 128x4096} & 1371.0 & 484.1 & 2.83 \\
Qwen3-0.6B & \texttt{b1: 4096x1024 @ 1024x1024} & 125.5 & 41.9 & 2.99 \\
Qwen3-0.6B & \texttt{b1: 4096x1024 @ 1024x3072} & 320.2 & 111.4 & 2.87 \\
Qwen3-0.6B & \texttt{b2: 4096x4096 @ 4096x128} & 177.6 & 53.7 & 3.31 \\
Qwen3-0.6B & \texttt{b1: 4096x1024 @ 1024x1536} & 168.8 & 63.7 & 2.65 \\
\midrule
Qwen3-1.7B & \texttt{b1: 4096x2048 @ 2048x3072} & 603.4 & 212.4 & 2.84 \\
Qwen3-1.7B & \texttt{b1: 4096x2048 @ 2048x6144} & 1191.1 & 402.7 & 2.96 \\
Qwen3-1.7B & \texttt{b1: 4096x3072 @ 3072x2048} & 617.2 & 208.8 & 2.96 \\
Qwen3-1.7B & \texttt{b8: 4096x4096 @ 4096x128} & 710.1 & 387.5 & 1.83 \\
Qwen3-1.7B & \texttt{b1: 4096x2048 @ 2048x1536} & 309.2 & 118.0 & 2.62 \\
Qwen3-1.7B & \texttt{b1: 4096x1536 @ 1536x2048} & 374.8 & 111.9 & 3.35 \\
Qwen3-1.7B & \texttt{b4: 4096x4096 @ 4096x128} & 284.5 & 134.9 & 2.11 \\
Qwen3-1.7B & \texttt{b1: 4096x2048 @ 2048x12288} & 2395.5 & 795.2 & 3.01 \\
Qwen3-1.7B & \texttt{b1: 4096x6144 @ 6144x2048} & 1111.7 & 412.1 & 2.70 \\
Qwen3-1.7B & \texttt{b1: 4096x2048 @ 2048x2048} & 470.8 & 147.4 & 3.19 \\
\bottomrule
\end{tabular}
\end{table*}

\paragraph{Operator and communication composition.}
For each paired configuration, durations are summed across ranks by operator
category or canonical communication primitive. Primitive names are normalized,
with send/receive variants grouped as point-to-point communication. Vectors
are aligned over the union of categories, filling absent entries with zero,
and normalized separately for operators and communication. Spearman
correlation and JSD are computed per configuration and summarized by their
medians.

Communication-primitive time shares have median rank correlations of
$0.949$ and $0.872$ and median JSD of $0.019$ and $0.035$; the dominant
primitive matches in 86.7\% and 72.7\% of configurations. Operator-category
correlations are $0.503$ and $0.531$, with median JSD of $0.436$ and $0.356$.
Table~\ref{tab:gpu-components} reports time shares pooled over the paired
configurations within each model and stack. The category breakdown makes
the implementation differences visible: on 910B3, \emph{other} and
\emph{shape\_transform} account for a large share of device time, while
\emph{attention} is a separate category only on the GPU stack. These
categories follow each profiler's classification and reflect each stack's
kernel composition.

\begin{table*}[t]
\centering
\small
\caption{Operator-category and communication-primitive time shares on GPUs and NPUs.}
\label{tab:gpu-components}
\begin{tabular}{lrrrr}
\toprule
Component & 0.6B GPU & 0.6B NPU & 1.7B GPU & 1.7B NPU \\
\midrule
\multicolumn{5}{l}{\textit{Operator categories}} \\
matmul & 28.6\% & 14.9\% & 38.9\% & 22.0\% \\
memory\_movement & 20.7\% & 4.2\% & 16.2\% & 3.3\% \\
elementwise & 18.6\% & 17.2\% & 19.1\% & 19.9\% \\
attention & 14.7\% & 0.0\% & 11.2\% & 0.0\% \\
memory\_init & 13.0\% & 0.2\% & 10.1\% & 0.2\% \\
other & 2.3\% & 40.3\% & 2.2\% & 34.1\% \\
shape\_transform & 1.6\% & 22.8\% & 1.4\% & 20.0\% \\
activation & 0.5\% & 0.2\% & 0.7\% & 0.3\% \\
embedding & 0.0\% & 0.0\% & 0.0\% & 0.0\% \\
optimizer & 0.0\% & 0.1\% & 0.1\% & 0.1\% \\
indexing\_or\_embedding & 0.0\% & 0.0\% & 0.0\% & 0.0\% \\
normalization & 0.0\% & 0.0\% & 0.0\% & 0.0\% \\
\midrule
\multicolumn{5}{l}{\textit{Communication primitives}} \\
all\_reduce & 54.4\% & 82.9\% & 54.3\% & 84.1\% \\
point\_to\_point & 42.3\% & 15.3\% & 41.3\% & 14.0\% \\
broadcast & 1.8\% & 0.9\% & 1.8\% & 0.7\% \\
reduce\_scatter & 1.5\% & 0.5\% & 2.7\% & 0.6\% \\
all\_gather & 0.0\% & 0.4\% & 0.0\% & 0.5\% \\
\bottomrule
\end{tabular}
\end{table*}

\paragraph{Communication overlap.}
Non-overlapped communication share is computed per rank as non-overlapped
communication time divided by stage time. Rank-level shares are summarized
by their median within each configuration, followed by the median over paired
configurations. The median non-overlapped communication share is 17.0\% and 10.9\% on V100 versus 28.5\%
and 30.0\% on 910B3, and the per-configuration correlation of this share across
stacks is only $\rho=0.179$ and $0.364$. Communication primitive ordering is
more consistent across stacks than the degree of communication overlap.

\clearpage
\end{document}